\documentclass[lettersize,journal]{IEEEtran}
\usepackage{amsmath,amsfonts}
\usepackage{algorithmic}
\usepackage{algorithm}
\usepackage{array}
\usepackage[caption=false,font=normalsize,labelfont=sf,textfont=sf]{subfig}
\usepackage{textcomp}
\usepackage{stfloats}
\usepackage{url}
\usepackage{verbatim}
\usepackage{graphicx}
\usepackage{hyperref}
\usepackage{cleveref}

\usepackage[numbers]{natbib}
\usepackage{booktabs}
\newcommand{\Real}{\mathbb{R}}

\begin{document}

\title{Neural Operator: Is data all you need to model the world? An insight into the paradigm of data-driven scientific ML}
\markboth{IEEE Transactions on Pattern Analysis and Machine Intelligence}%
{}

\author{Hrishikesh~Viswanath,
        Md~Ashiqur~Rahman,
        Abhijeet~Vyas,
        Andrey~Shor,
        Beatriz~Medeiros,
        Stephanie~Hernandez,
        Suhas~Eswarappa~Prameela,
        and~Aniket~Bera
\thanks{*H. Viswanath is the corresponding author.}%
\thanks{H. Viswanath, M. A. Rahman, A. Vyas, A. Shor, and A. Bera are with the Department of Computer Science, Purdue University, West Lafayette, IN, USA (e-mail: hviswan@purdue.edu).}%
\thanks{B. Medeiros and S. Hernandez are with the Hopkins Extreme Materials Institute, Johns Hopkins University, Baltimore, MD, USA.}%
\thanks{S. Eswarappa . Prameela is with the Department of Materials Science and Engineering, Department of Metallurgical Engineering and Department of Mechanical Engineering at the University of Utah, UT, USA}}

\maketitle

\begin{abstract}
Numerical approximations of partial differential equations (PDEs) are routinely employed to formulate the solution of physics, engineering, and mathematical problems involving functions of several variables, such as the propagation of heat or sound, fluid flow, elasticity, electrostatics, electrodynamics, and more. While this has led to solving many complex phenomena, there are some limitations. Conventional approaches such as Finite Element Methods (FEMs) and Finite Difference Methods (FDMs) require considerable time and are computationally expensive. In contrast, data-driven machine learning-based methods, such as neural networks, provide a faster, fairly accurate alternative, and, in particular, focus on neural operators, which have certain advantages such as discretization invariance and resolution invariance. This article aims to provide a comprehensive insight into how data-driven approaches can complement conventional techniques to solve engineering and physics problems, while also noting some of the open problems of machine learning-based approaches. We will note how these new computational approaches can bring immense advantages in tackling many problems in fundamental and applied physics.  
\end{abstract}


\section{Introduction}



\label{sec:Introduction}


Partial differential equations (PDEs) are an integral tool in mathematically modeling the physical world. They allow one to describe how a quantity changes with respect to multiple variables and have allowed physicists to model various phenomena in fluid flow, electrodynamics, and quantum mechanics. An example family of generic PDEs can be represented as shown in equation \ref{eq:1},
\begin{align}\label{eq:1}
    (L_au)(x) = f(x),~~~x \in D,\\
u(x) = 0,~~~x \in \delta D \nonumber
\end{align}
for some $a \in A, f\in L$, where $A,L$ are Banach spaces, $D$ is the domain of the PDE and $u : D \rightarrow \mathbf{R},u\in U $ is the solution function.
While PDEs are all around us, it is oftentimes very difficult for one to solve them analytically. The best that one can achieve is an approximation of the true solution of the PDE. The most popular approaches to solving PDEs are numerical methods such as finite difference methods (FDMs) \cite{godunov1959finite}, finite element methods (FEMs) \cite{zienkiewicz2005finite}, and finite volume methods (FVMs) \cite{eymard2000finite} as they are able to approximate solutions to PDEs with high amounts of accuracy. However, they are computationally expensive.   



Numerical methods have traditionally been used to solve PDEs, but in an effort to reduce the computational cost and obtain the solutions more quickly, researchers are exploring data-driven approaches to approximate the solutions to PDEs. These methods come under the overarching paradigm of machine learning approaches, which utilize data-driven algorithms that allow a program to learn and improve from experience.
 Recent advances in deep learning have allowed researchers to develop neural network architectures and training strategies to model and approximate scientific problems, many of which are modeled with PDEs. This has led to the rise of a new field called Scientific Machine Learning or AI for science.
 
Deep neural networks have been applied to a multitude of problems in material physics, thermodynamics, and fluid dynamics problems \cite{cai2022physics}, \cite{mao2020physics}, \cite{cai2021physics}, \cite{thais2022graph} due to their innate ability to learn complex relationships between physical entities. In particular, the foundation of this paradigm was built on physics-informed neural networks (PINNs) \cite{raissi2019physics} and operator-based networks such as Deep-ONet \cite{lu2019deeponet} and Fourier Neural Operator \cite{li2020fourier}. While PINNs incorporate PDE residuals into their training loss, neural operators are primarily data-driven and do not require access to explicit forms of the PDEs. 

While neural networks are able to approximate any function, which is a map between finite-dimensional spaces \cite{cybenko1989approximation}, to approximate an operator, which is a map between infinite-dimensional spaces, a network of infinite length is required \cite{guss2019universal}. A neural operator is a generalization of a neural network that maps between infinite-dimensional spaces \cite{kovachki2021neural}. These operators are capable of approximating highly nonlinear solution operators of PDEs. Furthermore, neural networks trained as 
neural operators are discretization-invariant and up to $\approx$1000x faster than typical neural networks in approximating the solution of PDEs, as shown by \cite{li2020fourier}. Current state-of-the-art neural operator architecture includes the DeepONet, graph neural operator (GNO) \cite{li2020multipole, li2024geometry}, Fourier neural operator (FNO) \cite{li2020fourier} and its variants, geo-Fourier neural operator (Geo-FNO) \cite{li2022fourier}, and physics-informed neural operator (PINO) \cite{rosofsky2022applications}. More recently, transformer-based architectures and diffusion-based architectures have been designed to work as operators. These include, AFNO \cite{guibas2021adaptive}, GNOT \cite{hao2023gnot}, IPOT \cite{lee2024inducing}, and diffusion frameworks such as \citep{zheng2023fast, haitsiukevich2024diffusion, oommen2024integrating}. Extending these further, architectures such as CoDANO \cite{rahman2024pretraining} place themselves in the realm of foundational models for PDEs, a class of pretrained architectures applicable for a broad range of benchmark problems. Meanwhile, UPT \cite{alkin2024universal} is a class of modular architectures with detachable components for large-scale problems.  
Despite these successes, significant challenges remain. These models often require high-fidelity datasets to prevent overfitting. Zero-shot generalization across PDE parameterizations \cite{viswanathgradient}, geometries \cite{li2024geometry}, and scalability to extremely high resolution settings \cite{xu2025amr} remain open problems. 

\begin{figure*}[h]
    \centering
    \includegraphics[width=1\textwidth]{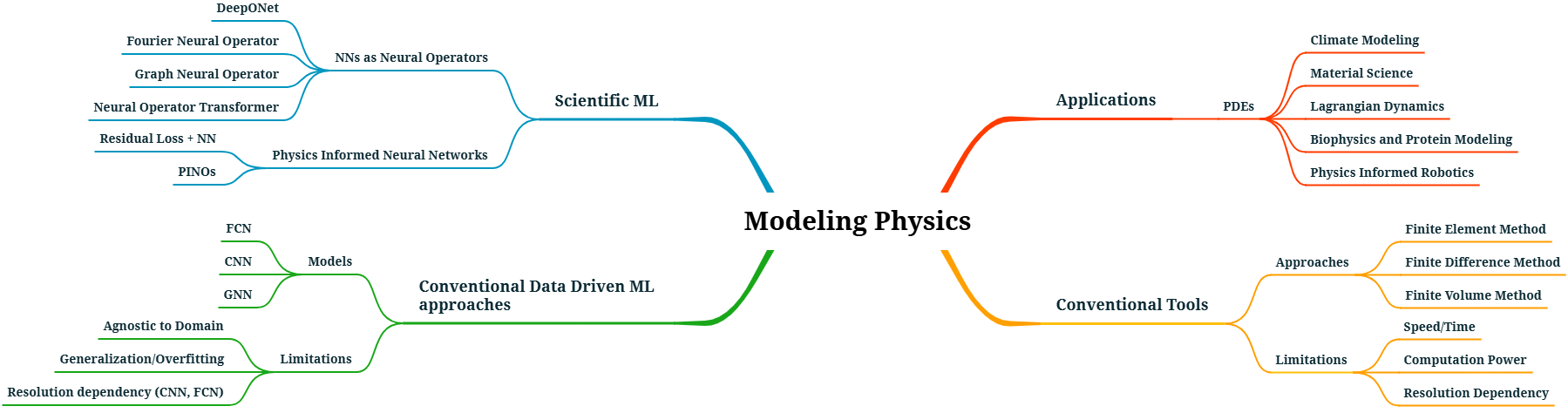}
    \caption{The above figure represents the logical flow of how PDEs can be solved using various methods, highlighting existing ML techniques and various families of neural operator-based techniques}
    \label{fig:Pipeline}
\end{figure*}

\paragraph{Scope and contributions} This survey aims to provide a comprehensive overview of data-driven and physics-informed neural operators in the broader realm of scientific ML. We evaluate these models along three critical axes: scalability, generalizability, and sensitivity to training data quality, and discuss open problems in each of these fields. Figure \ref{fig:Pipeline} provides an overview of both conventional and machine learning-based approaches to solving PDEs. We will note how these new computational approaches can bring immense advantages in tackling many open problems in fundamental and applied physics.  

\textit{Organization:} The review is organized as follows. In \cref{sec:ML_app}, we first discuss conventional machine learning approaches for solving PDE-based problems and provide a discussion of PDE-based physics-informed losses. We then discuss the neural operator family of architectures in \cref{sec:NO} and more advanced variants in \cref{sec:advanced_no}, providing an overview of data-driven learning paradigms across the three axes, followed by a discussion on open problems in these areas in \cref{sec:Open_prob}. In \cref{sec:standardization}, we discuss existing benchmark datasets and controlled settings for addressing these problems, and in the subsequent \cref{sec:app}, we expand to how these strategies can be applied to three classes of problems - large scale forecasting \& long-range temporal problems, modeling heterogeneous systems, and lastly, inverse problems. In the concluding sections, we discuss the realization of these architectures in real-world settings and their adaptability to this class of problems. 

\section{ML-based approaches}
\label{sec:ML_app}
Neural Network-based approaches to solve PDEs are broadly categorized into data-driven and physics-informed. In the former case, the computational domain is discretized, and the function values are represented as finite-dimensional tensors (e.g., matrices for 2D grids). The ground truth solution is pre-computed on the discretized mesh, and the neural network, serving as the surrogate, is trained to approximate these function mappings by minimizing a supervised loss objective. Conversely, the latter approach embeds physical laws directly into the loss objective by defining it in terms of the PDE residual, thus avoiding the need to pre-compute training datasets.

These approaches have been shown to perform quite well in parametric PDE approximations and high-dimensional PDEs. Once trained, these neural surrogates enable solving the PDE with different initial and boundary conditions \cite{karniadakis2021physics}. 

While these techniques offer promising results, in most situations, they do not outperform higher-order classical numerical methods \cite{jentzen2023algorithmically}, but rather, provide a way for faster approximation of the solutions. Moreover, these networks are limited by the resolution of the mesh on which they are trained. To that end, operator learning was proposed to learn mesh invariant solutions to PDEs \cite{chen2023deep}. Neural operators, therefore, refer to neural networks that are trained to learn infinite-dimensional function mappings in a discretization-invariant manner. 

The following subsections discuss various neural network-based architectures, and Table \ref{tab:differences_classical_dd} highlights the differences between classical solvers and data-driven approaches. 

\subsection{Standard neural architectures and their limitations}
Prior to the development of operator learning, research focused on leveraging classical deep learning architectures such as fully connected networks (FCNs) and convolutional neural networks (CNNs) to approximate PDE solutions. 

\paragraph{Fully connected neural networks}
Early approaches utilized FCNs as universal function approximators \cite{hornik1989multilayer} to represent the solution as a continuous function of space and time. The seminal work by \cite{lagaris1998artificial} introduced an FCN-based approach to approximate pointwise solutions. While this yielded differentiable closed-form solutions in lower dimensions, it struggled to generalize to higher dimensions. This was later addressed in the deep Galerkin method (DGM) \cite{sirignano2018dgm}, which leveraged Monte-Carlo sampling to sample training points. This approach was mesh-free but remains sensitive to sampling quality and optimization stability. Another similar class of methods is the deep Ritz method (DRM) \cite{yu2018deep}, which uses a variational energy functional as the loss objective. Comparisons between these paradigms suggest that while DGM generally performs well for smooth solutions, DRM offers advantages for solutions with lower regularity. However, recent studies indicate that this distinction is not absolute; DRM can outperform DGM in high-dimensional settings even for smooth solutions, while DGM remains competitive for specific low-regularity cases depending on the sampling strategy \cite{chen2020comparison}. Furthermore, the handling of boundary conditions plays a critical role in the success of both architectures. While soft-constraints (penalty terms) are flexible, enforcing boundary conditions exactly,when analytically feasible,has been shown to significantly stabilize training and improve approximation accuracy for both DGM and DRM frameworks.

\paragraph{Convolutional neural networks}
 To better capture local spatial correlations, convolutional neural networks (CNNs) have been adopted as efficient alternatives to FCNs. Early applications demonstrated CNNs' capacity to approximate elliptic PDEs by learning spatial features effectively \cite{o2015introduction}. Notable frameworks include ConvPDE-UQ \cite{winovich2019convpde}, which utilizes Green's functions to construct lightweight solvers that approximate solutions in a single forward pass. While offering significant computational efficiency over FEM, this approach struggles with inhomogeneous systems and mixed boundary conditions. To address geometric complexity, PhyGeoNet \cite{gao2021phygeonet} employs an elliptic coordinate mapping to transform irregular physical domains into regular reference grids, enabling the solution of parametric PDEs without labeled data. However, PhyGeoNet remains limited to steady-state problems.

For spatiotemporal dynamics, hybrid architectures have emerged. PhyCRNet \cite{ren2022phycrnet} and recent ConvLSTM-based models \cite{MAVI2023115944} combine convolutional layers for low-dimensional spatial feature extraction with recurrent units for temporal evolution. While these hybrid models improve temporal extrapolation, they generally remain constrained to fixed spatial discretizations and often fail to generalize well across diverse initial conditions.

\subsection{Making neural networks physics-informed}

The architectures described above can be trained using the physics-informed paradigm, which embeds governing physical laws directly into the loss function to minimize reliance on pre-computed data. This paradigm is generally categorized by the formulation of the physical constraint: residual-based or variational.

The standard physics-informed neural network (PINN) and the deep Galerkin method (DGM) both follow the residual-based approach. The training objective is defined as the weighted sum of the PDE residuals, boundary conditions, and initial conditions \cite{cuomo2022scientific}. While standard PINNs typically employ fixed or adaptive collocation grids, DGM specifically leverages Monte-Carlo sampling to estimate these residuals, enabling scalability to high-dimensional spaces.

In contrast, the deep Ritz method (DRM) \cite{yu2018deep} enforces physics constraints through a variational energy minimization approach. Instead of pointwise residuals, DRM minimizes the energy functional of the PDE, requiring lower solution regularity. Regardless of the specific formulation, these physics-informed strategies fundamentally solve an optimization problem for a specific PDE instance, ensuring validity without requiring ground-truth supervision.

However, these methods struggle with optimization. To evaluate the PDE residual, the network must explicitly calculate the derivatives of its output with respect to the input coordinates. This process relies on automatic differentiation operation (i.e. \texttt{autograd}), which becomes computationally expensive for high-order equations and often leads to severe training instabilities \cite{viswanathgradient}. To overcome these bottlenecks, recent works have proposed weakly supervised optimization strategies, specifically for linear elliptic PDEs \cite{nam2024solving, viswanathgradient}. These approaches leverage the Feynman-Kac formalism to generate weak high variance supervision signals. Instead of minimizing a residual based on derivatives, the network is trained to regress to stochastic solution estimates obtained via random walks. This effectively bypasses the stability issues of automatic differentiation by utilizing the neural network's inherent ability to reduce variance and smooth the solution during training.

Nevertheless, whether using standard PINNs or stochastic variants, a fundamental limitation remains: these are single-instance solvers. A change in PDE parameterization requires the model to be retrained from scratch. This motivates the development of neural operators, which aim to learn the mapping for an entire family of PDEs. 

\begin{table}[h]
    \centering
    \resizebox{8cm}{!}{
    \begin{tabular}{cc}
        \toprule
        \textbf{Conventional Methods} &  \textbf{Scientific ML Methods}\\
        \midrule
        Slower on Fine grids & Slow training time, Fast \\& Inference Time \\
        & \\
        Discretization Dependent & Resolution \\ & \& discretization invariant \\ & (Neural Operators)\\
        & \\

        Requires explicit form & Only requires training data\\& (+ PDE Loss in case of PINNs) \\
        & \\
        Higher computation power  & Once trained, comparable \\
        for larger meshes & computation power \\
        & for different resolutions \\
        \bottomrule
        
    \end{tabular}}
    \caption{This table highlights the main differences between conventional methods and data-driven methods}
    \label{tab:differences_classical_dd}
\end{table}

\section{Neural Networks for Operator Learning}
\label{sec:NO}

To overcome the resolution-dependency of standard neural networks, operator learning proposes a paradigm shift: instead of learning the solution to a specific discretized instance, the model learns the operator $\mathcal{G}: \mathcal{A} \to \mathcal{U}$ mapping between infinite-dimensional function spaces.
The defining characteristic of a neural operator is discretization invariance: the network is trained on a finite collection of input-output pairs $\{a_i, u_i\}_{i=1}^N$ (e.g., initial condition to solution) at a specific resolution, but can be evaluated at any arbitrary resolution during inference. This capability, often termed ``zero-shot super-resolution," allows for:
\begin{enumerate}
\item Rapid inference: Approximating solutions orders of magnitude faster than classical solvers.
\item Mesh independence: Training on coarse grids (e.g., $40 \times 40$) and evaluating on fine grids (e.g., $256 \times 256$) without retraining.
\end{enumerate}

\textbf{Taxonomy of architectures:} The field of operator learning originated with two primary architectural paradigms: branch-trunk architectures (e.g., DeepONet) and integral kernel architectures (e.g., FNO, GNO). These foundational models were soon followed by task-specific improvements aimed at generalizing to irregular geometries and non-uniform meshes (Geo-FNO), enhancing computational efficiency through optimized latent representations (GNOT, Multiwavelet), improving expressivity via modern transformer-based and generative backbones (AFNO, GNOT, Diffusion Operators), building pre-trained foundational models (CoDANo),  modularizing the architecture for adaptability (UPT), and lastly, incorporating physical constraints to improve generalizability (physics-informed neural operators PINO). The following subsections review these developments. In \cref{tab:Summary}, we provide a comparative summary of seminal neural operator architectures, highlighting their key characteristics, advantages, limitations, and initial application domains. 

\subsection{Branch-trunk architecture: DeepONet} The deep operator network (DeepONet) \cite{lu2019deeponet} is formulated based on the universal approximation theorem for operators \cite{chen1995universal}. It explicitly decomposes the operator learning task into two sub-networks: a branch net that encodes the input function $a(x)$ (discretized at fixed sensor locations) and a trunk net that encodes the continuous query coordinates $y$.  The final output is computed as the dot product of these two latent representations. This architecture has demonstrated broad versatility, effectively modeling dynamic systems in material physics \cite{goswami2022physics} and biological transport in aortic dissections \cite{yin2022simulating}. To address the limitation of fixed input sensors, extensions such as the Variable-Input Deep Operator Network (VIDON) \cite{prasthofer2022variable} have been proposed to handle inputs defined on variable discretizations. Similarly, to mitigate the spectral bias inherent in standard neural networks, multi-scale DeepONet \cite{liu2021multiscale} utilizes frequency-scaling transformations. This allows the network to approximate high-frequency oscillatory functions (e.g., seismic excitations) by mapping them onto lower-frequency manifolds, a capability often lacking in standard FCN-based trunks.

\subsection{Integral kernel architectures: FNO and GNO}A second family of architectures formulates the operator learning problem as an iterative integral kernel transformation: $(K_a u)(x) = \int_D \kappa(x, y, a(x), a(y)) u(y) dy$. Graph neural operators (GNO) \cite{li2020neural} approximate this integral via message passing on graph-discretized domains. By treating mesh points as nodes and kernel integration as neighborhood aggregation, GNOs naturally handle unstructured grids. Perhaps the most prominent realization of this paradigm is the Fourier neural operator (FNO) \cite{li2020fourier}, which parameterizes the integral kernel in the frequency domain.  By applying the fast Fourier transform (FFT), the global convolution operation becomes a simple linear multiplication in spectral space. By truncating the Fourier series at a maximal mode $k_{max}$, the FNO achieves a resolution-invariant parameterization that is highly efficient for problems with periodic boundary conditions or smooth global features. 

\section{Advanced and task-specific neural architectures}
\label{sec:advanced_no}
While the foundational architectures (DeepONet, FNO, GNO) established the operator learning paradigm, they are often constrained by specific assumptions, such as the requirement for rectangular domains in FNO or the need for fixed sensors in DeepONet. To address these limitations, the field has evolved toward specialized architectures tailored for complex physical constraints. The following section discusses these advanced variations, categorizing them by their primary contributions. Table \ref{tab:tab_links} provides links to the source code for these various frameworks.

\textbf{Geometry-aware architectures: } Standard FNOs are limited to rectangular domains due to their reliance on the FFT. To extend operator learning to complex, irregular geometries, several variants have been proposed. Geo-FNO \cite{li2022fourier} addresses this by learning a coordinate deformation mapping $x \to \xi$ that transforms the irregular physical domain into a uniform latent lattice where FFTs can be applied. Alternatively, Geometry-informed neural operator (GINO) \cite{li2024geometry} adopts a hybrid approach, stacking an FNO between graph neural operator (GNO) encoders and decoders. This allows the model to process irregular geometries via the GNO layers while retaining the spectral efficiency of the FNO in the latent space.

\textbf{Transformer and generative variants} Recent work has integrated modern deep learning primitives into the operator framework to enhance scalability and expressivity. Architectures like AFNO \cite{guibas2021adaptive}, IPOT \cite{lee2024inducing} and GNOT \cite{hao2023gnot} interpret the discretized field as a sequence of tokens, utilizing transformer attention mechanisms to capture global dependencies more flexibly than fixed spectral modes. Transolver is another architecture that proposes physics attention, derived from the integral kernel transform. Moreover, AFNO is designed to work in image space, treating image inputs as a sequence of tokens. 

To capture stochasticity or multimodal solutions, generative frameworks have also been adapted. GANO \cite{rahman2022generative} utilizes adversarial training to learn operator distributions, while DSNO \cite{zheng2023fast} leverages diffusion models in the spectral domain to generate continuous solution trajectories from Gaussian noise.

\textbf{Physics-informed neural operator (PINO)} While the architectures above are primarily data-driven, the physics-informed neural operator (PINO) \cite{li2021physics} creates a hybrid paradigm. PINO augments the operator loss with a PDE-residual constraint, similar to PINNs. Crucially, unlike PINNs, which optimize a single solution, PINO uses the physics loss to regularize the learning of the operator itself. This improves generalization on small datasets and ensures physical consistency (e.g., mass conservation) while retaining the fast inference speeds of standard operators.

\textbf{Multiwavelet Neural Operator (MWNO)} While FNOs rely on the global Fourier basis, which can struggle with local discontinuities, the multiwavelet neural operator \cite{gupta2021multiwavelet} proposes decomposing the integral kernel using an orthonormal wavelet basis. By embedding inverse wavelet filters, the operator projects the kernel into multiwavelet polynomial bases, exploiting orthogonality and vanishing moments to achieve highly compact, sparse representations of the operator. Empirically, this approach has demonstrated superior performance on systems with localized high-frequency features (e.g., 2D Navier-Stokes velocity fields) and shows promise in super-resolution tasks. However, it faces challenges in generalizing from low-frequency training data to high-frequency test signals compared to the global receptive field of the FNO.

\textbf{Foundational and modular frameworks} The most recent frontier in operator learning focuses on developing \textit{Foundational Models} style architectures designed to generalize across broad classes of PDEs rather than specific instances.

\textit{Foundational pre-training} Architectures such as CoDANO (Codomain Attention Neural Operator) \cite{rahman2024pretraining} aim to establish this foundational capability through pre-training. By introducing codomain attention, the model learns to attend to the correlations between output physical variables, allowing it to act as a general-purpose solver applicable to a wide range of benchmark problems without extensive retraining.

\textit{Modular architectures} Concurrently, the Universal Physics Transformer (UPT) \cite{alkin2024universal} framework addresses the challenge of large-scale multi-physics problems through modularity. UPT employs an Encode-Process-Decode paradigm where components are detachable and reusable. This modular design allows different encoders (e.g., for different geometries or boundary conditions) to be paired with a shared latent processor, facilitating scalable multi-task learning across diverse physical domains.

\section{Open research problems in Data Driven Scientific ML}
\label{sec:Open_prob}
\label{sec:standardization}
\begin{figure*}
    \centering
    \includegraphics[width=0.9\textwidth]{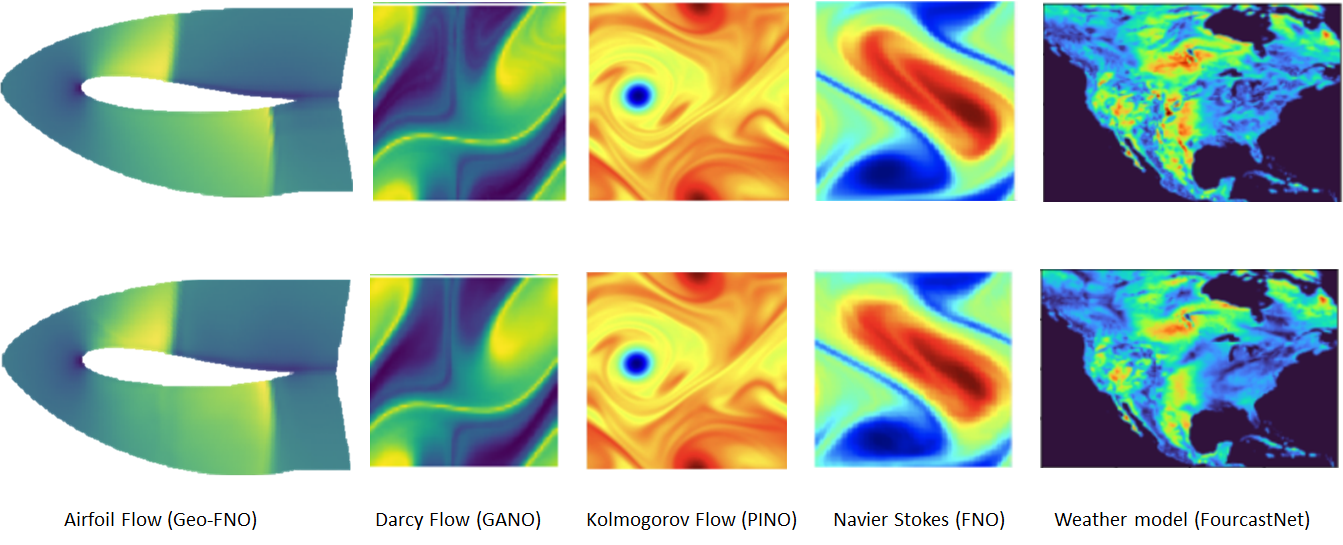}
    \caption{The above figure highlights the performance of various neural operator based architectures in solving different types of problems. It provides a visual comparison between generated solution an ground truth. The images on the top are the ground truth values while the ones on the bottom row are generated by the operator models. The first image on the left represents the airfoil Flow simulation generated by the geo-FNO architecture \cite{liu2022learning}. The second one is the Darcy Flow simulation generated by the GANO architecture \cite{rahman2022generative}. The third one represents the Kolmogorov Flow generated by the PINO architecture \cite{li2021physics}, the fourth pair represents the Navier-Stokes equations simulated by the vanilla FNO \cite{li2020fourier} and the last one is the weather forecast model generated by the FourcastNet \cite{pathak2022fourcastnet}.}
    \label{fig:PDESummary}
\end{figure*}
While Neural Operators and Physics-Informed frameworks offer a paradigm shift from instance-based solving to operator learning, they are not yet a panacea. The transition from theoretical universality to practical deployment poses significant challenges. We categorize these open problems along three critical axes: data sensitivity, generalizability, and scalability.

\subsection{Data sensitivity and robustness} Unlike classical numerical solvers (FEM/FDM), where error bounds are determined by mesh refinement, neural operators are sensitive to training setups. 

\paragraph{Grid sensitivity} Theoretically, neural operators are discretization-invariant. However, in practice, models like the FNO are susceptible to aliasing errors when the inference grid differs significantly from the training grid. As shown by \cite{fanaskov2022spectral}, high-frequency errors can propagate through the network, causing outputs to deviate from the ground truth upon mesh refinement. 

\paragraph{Noise accumulation} For long-horizon temporal inferences, auto-regressive operators suffer from error compounding. Since the network learns a mapping between consecutive time steps, small approximation errors in early steps accumulate, leading to significant divergence in non-linear temporal evolutions \cite{sanchez2020learning}. This is often addressed by learning to predict a window of time frames, adding random-walk noise for robustness or through strategies such as teacher forcing \cite{toomarian1992learning}. 

\paragraph{Data requirements} Data-driven operators require high-fidelity datasets to cover the space of the PDE. In regimes where data is scarce or noisy, models often converge to spurious local minima that fail to respect physical laws \cite{zhu2023reliable}. 

\subsection{Generalizability and optimization stability} A core promise of operator learning is generalization across parameters and geometries. However, attaining this in complex physical regimes remains difficult. 

\paragraph{Optimization issues in physics-informed neural operators} When enforcing physics constraints (as in PINO), the optimization landscape often becomes unstable. \cite{krishnapriyan2021characterizing} demonstrated that in problems with high convection coefficients or stiff reaction-diffusion terms, physics-based losses fail to guide the optimizer to the correct solution. This is often due to the spectral bias of neural networks, which struggle to capture high-frequency discontinuities or sharp shock waves without specialized regularization. 
\paragraph{Geometric generalization} While architectures like Geo-FNO and GINO address irregular meshes, generalizing to unseen large-scale topologies of the order of millions of points/cells (e.g., training on airfoils and testing on turbine blades) remains an open challenge. Most current models effectively interpolate within a fixed geometric distribution but struggle to extrapolate to topologically distinct domains \cite{chen2022crom, ma2023learning}.

\subsection{Scalability and computational trade-offs} The third axis concerns the computational cost of scaling these architectures to high-dimensional, industrial-grade problems. 
\paragraph{The parameterization trade-off} Recent studies on over-parameterization \cite{kontolati2023influence} suggest that larger models generally yield lower generalization errors and that DeepONet variants maintain robustness regardless of parameter count. However, this comes at a steep training cost. 

\paragraph{Training vs. inference} There is a fundamental trade-off between offline training and online inference. \cite{grossmann2023can} noted that while PINNs and operators can be slower than FEM during the training/optimization phase, they offer orders-of-magnitude speedups during the inference phase. The challenge lies in reducing the break-even point: making the training efficient enough to justify the switch from classical solvers for problems that do not require real-time inference. 

\paragraph{Curse of dimensionality} Despite the success of DeepONet and FNO in 2D/3D space, scaling to high-dimensional PDEs (e.g., Boltzmann equations) remains computationally prohibitive for grid-based operator methods, necessitating further research \cite{mandl2025separable}.

The impact of these open challenges varies across scientific domains. For instance, stability and long-term error accumulation are crucial in climate modeling, whereas material science is more constrained by the need to generalize across irregular, heterogeneous geometries \cite{liu2022learning}. To reflect this, the following section organizes existing literature by physical regimes, highlighting specific architectural choices for each class of problem.

\section{Standardizing Operator Research with Canonical Benchmarks}

Academic studies in operator learning often require a set of canonical PDE problems and standardized datasets that stress-test specific capabilities. In this section, we discuss the benchmark and evaluation settings for both academic and large-scale problems. \Cref{fig:PDESummary} presents an overview of canonical problems often used to benchmark operator architectures. 

\subsection{Canonical problem setups}
\begin{figure}[th!]
    \centering
    \includegraphics[width=1\linewidth]{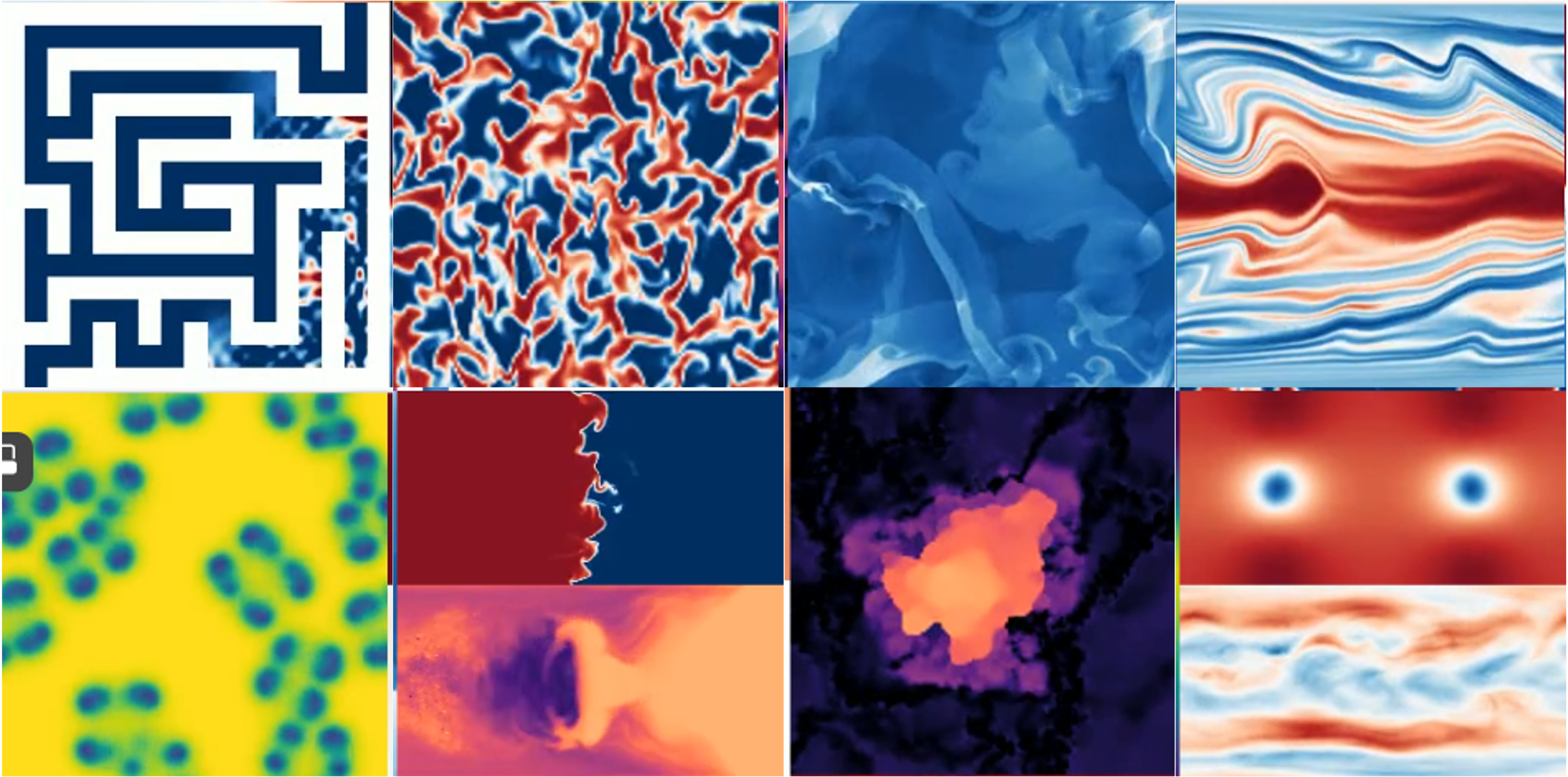}
    \caption{Simulating fluid-flow over complex settings is a challenging task. The above illustration, presented in the Well \cite{ohana2024well}, highlights the nature of this problem.}
    \label{fig:flowbench}
\end{figure}

\textbf{Eulerian regimes} Eulerian benchmarks evaluate an operator's ability to map fields on Eulerian frame of reference. 
\paragraph{Steady-state and linear problems} The Darcy Flow equation (porous media) serves as the standard test for mapping static, irregular coefficients to smooth pressure fields. Similarly, the Poisson and heat equations are frequently used as baselines to verify mesh-invariance and convergence rates on simple diffusive dynamics. 
\paragraph {Non-linear dynamics and discontinuities} The 1D viscous Burgers' equation is the primary testbed for shock formation, evaluating robustness to sharp discontinuities. For chaotic dynamics, the incompressible Navier-Stokes equations are employed in specific configurations: the Kolmogorov Flow (periodic boundary conditions with sinusoidal forcing) tests stability over long temporal rollouts, while the Karman Vortex Street (flow past a cylinder) evaluates the modeling of wake dynamics and vortex shedding frequencies. 

\begin{figure}[th!]
\centering
\subfloat[The GNS dataset \cite{sanchez2020learning} representing controlled behaviors of fluid substances obtained using MPM and SPH based simulation models]{%
  \includegraphics[width=0.4\textwidth]{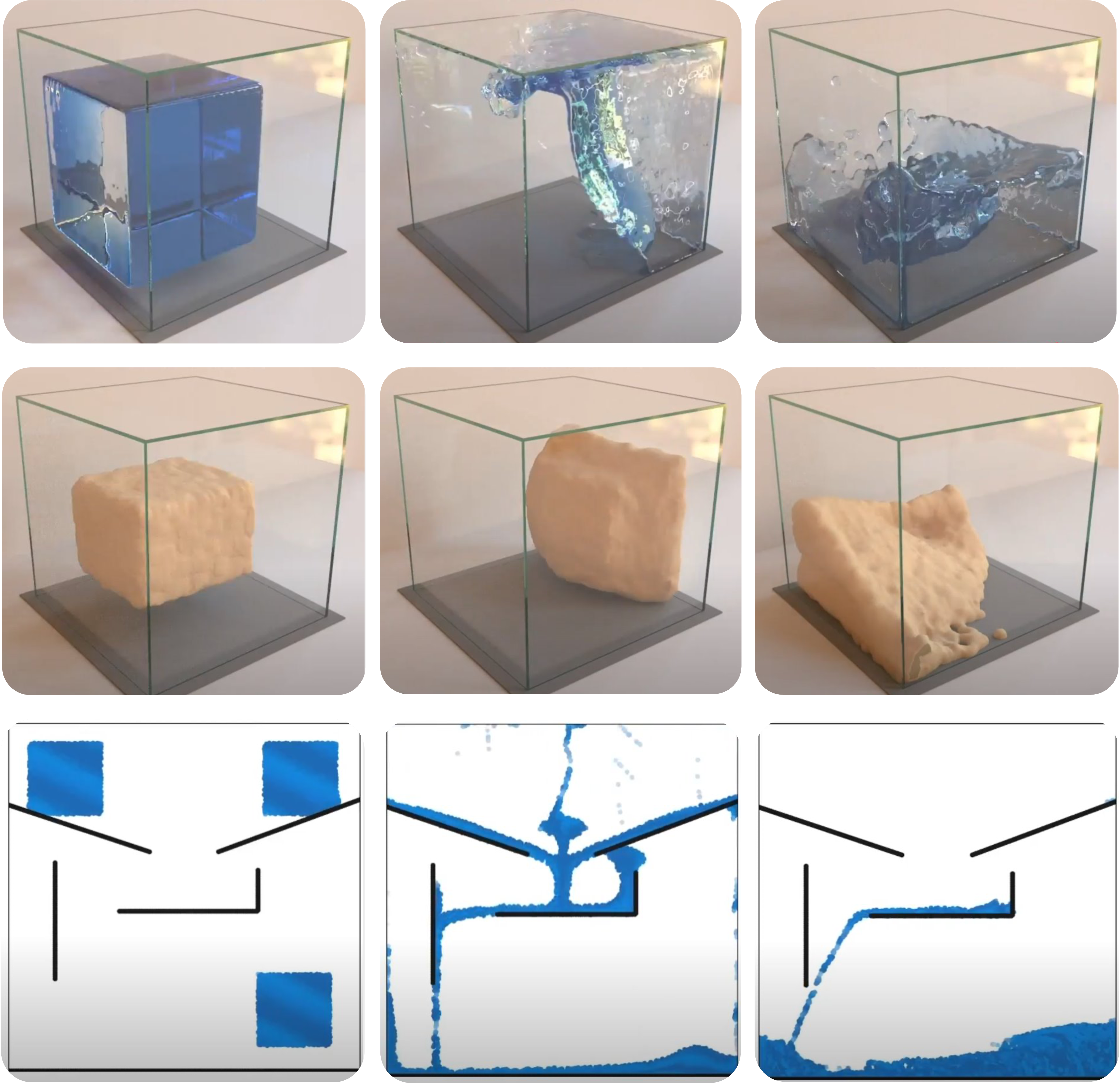}%
  \label{fig:gns}%
}
\hfil
\subfloat[Eulerian (left) and Lagrangian and  mesh based simulation datasets from MeshGraphNet \cite{pfaff2020learning}]{%
  \includegraphics[width=0.4\textwidth]{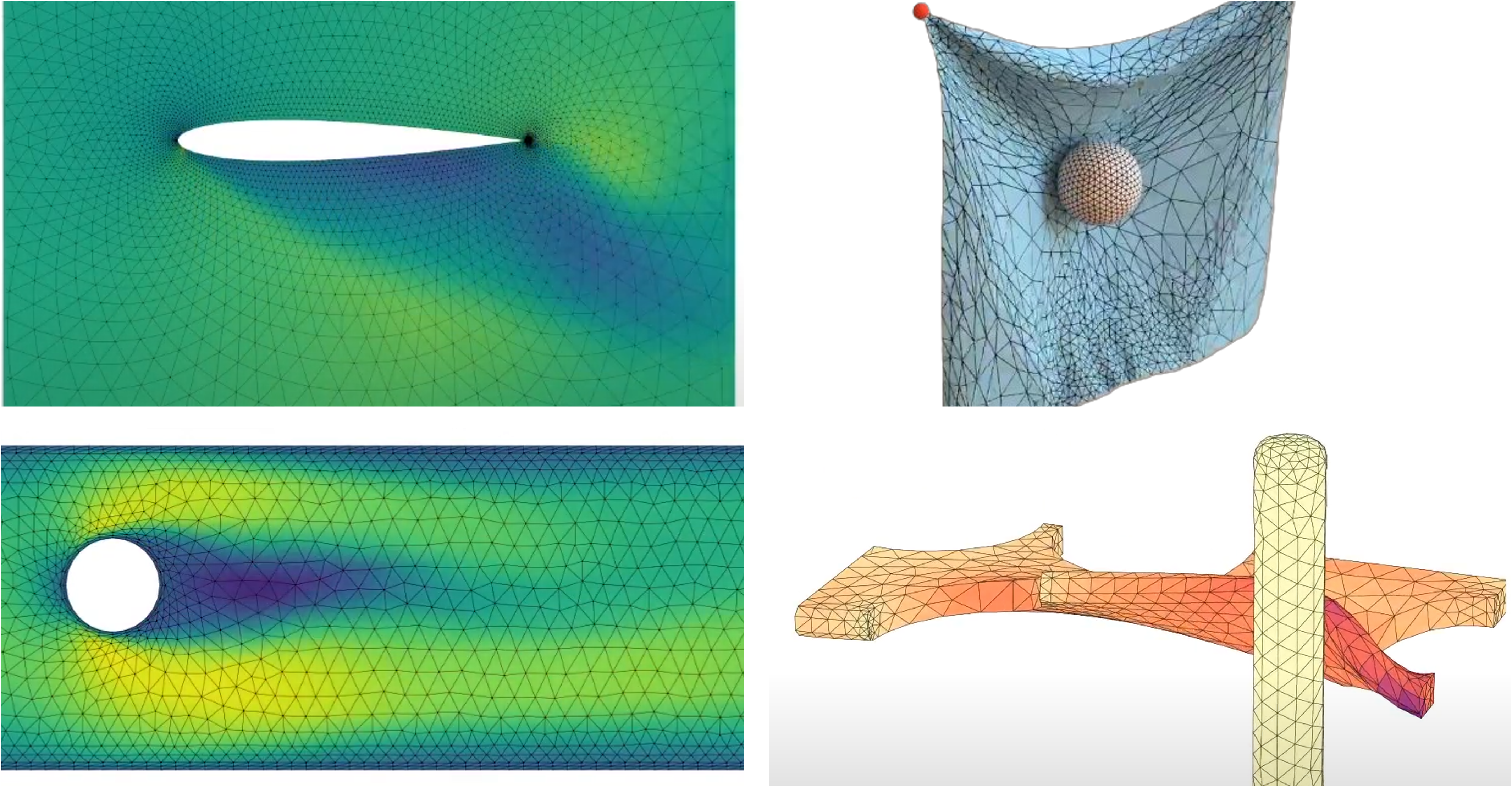}%
  \label{fig:meshgraphnet}%
}
\caption{The above datasets are commonly used in data-driven Lagrangian settings, often to benchmark models in computer graphics tasks for simulating fluids, cloth etc.}
\label{fig:gnns}
\end{figure}

\textbf{Lagrangian regimes} Lagrangian benchmarks evaluate the operator's capacity to model physical interactions on a dynamic, unstructured graph. These tasks bridge Scientific ML and computer graphics, focusing on scenarios involving large deformations and complex contact dynamics. Prominent benchmarks include the datasets introduced with the Graph Network Simulator (GNS) \cite{sanchez2020learning}, which require predicting the temporal evolution of granular materials (sand), fluids (SPH), and deformable solids (MPM). \Cref{fig:gnns} depicts what these datasets look like. Building on this, LagrangeBench \cite{toshev2023lagrangebench} offers a standardized suite of 2D and 3D fluid mechanics datasets, establishing rigorous baselines for particle-based learning.

\subsection{Benchmark datasets}

The evaluation of these canonical problems relies on datasets of varying geometric complexity, ranging from academic proofs-of-concept to industrial-grade scenarios.

\paragraph{Regular and parametric geometries} Foundational works typically utilize datasets generated via spectral solvers or finite element methods (e.g., FEniCS) on uniform Cartesian grids. To test generalization, these are often augmented with irregular parameterized geometries, such as airfoils or pipe bends, where the domain shape itself acts as an input parameter.

\paragraph{Industrial and complex geometries} To bridge the gap to real-world engineering, recent benchmarks utilize complex, non-smooth geometries. 

\textbf{General-purpose suites (PDEBench)} PDEBench \cite{takamoto2022pdebench} is currently the most extensive benchmark suite for time-dependent Scientific ML. It encompasses diverse physical systems, including compressible Navier-Stokes, diffusion-reaction, and shallow water equations, across varying dimensions (1D/2D/3D) and parameter regimes. Unlike earlier datasets, PDEBench provides standardized metrics for both single-step prediction and long-term autoregressive rollouts.

\textbf{Aerodynamics and geometry (AirfRANS \& Ahmed body)} To test geometric generalization, AirfRANS \cite{bonnet2022airfrans} offers high-fidelity Reynolds-averaged Navier-Stokes (RANS) simulations over diverse NACA airfoils, challenging operators to predict surface pressure and skin friction distributions accurately. For 3D turbulence, the Ahmed body dataset \cite{meile2011experiments} remains the standard for automotive aerodynamics, testing the resolution of wake structures on non-smooth industrial geometries.

\textbf{Planetary scale (WeatherBench 2)} For global forecasting models, WeatherBench 2 \cite{rasp2024weatherbench} serves as the definitive benchmark. It provides processed ERA5 reanalysis data and standardized evaluation metrics (RMSE, ACC) for medium-range weather forecasting, allowing direct comparison between data-driven operators (like FourCastNet) and operational numerical weather prediction (NWP) models.

\begin{figure}[ht!]
    \centering
    \includegraphics[width=1\linewidth]{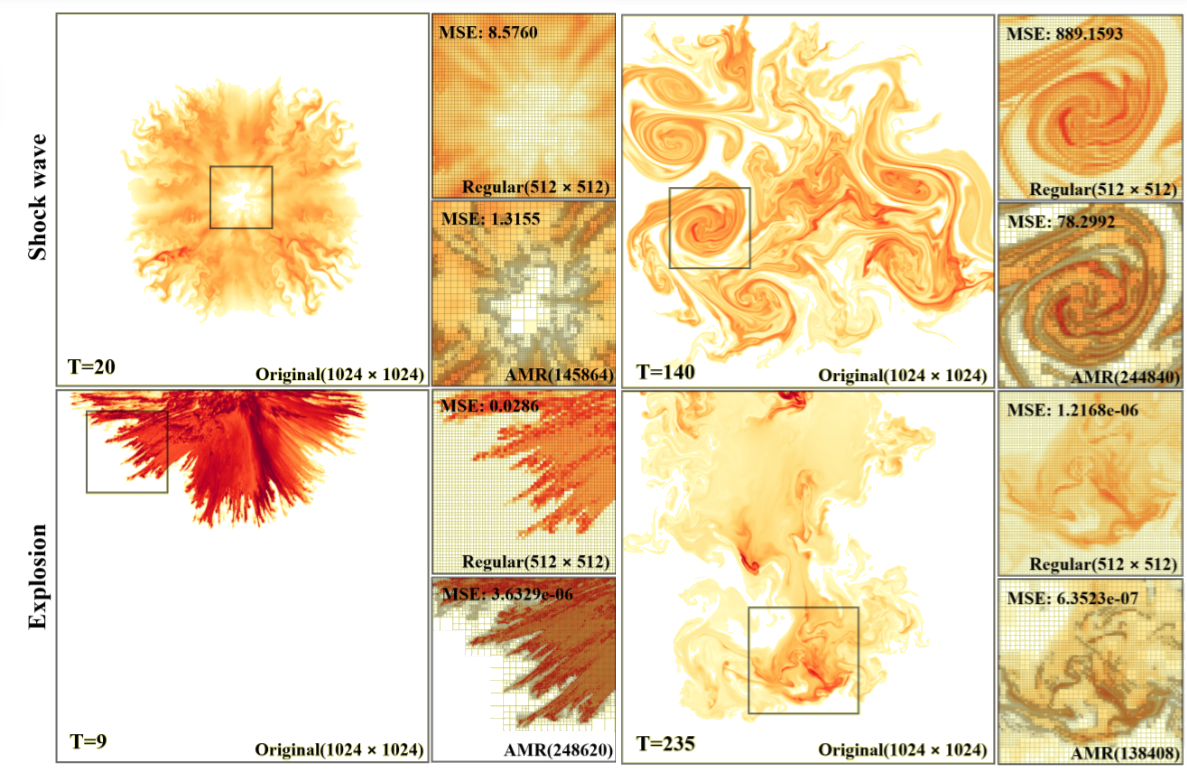}
    \caption{Forecasting behaviors in adaptive mesh problems is an open challenge. Particularly, in handling varying grid resolutions across time-steps. AMR Transformer \cite{xu2025amr}, proposes a data-driven adaptive octree based learning technique to model these class of problems. }
    \label{fig:amr}
\end{figure}

\textbf{Multi-resolution and adaptive meshes} To validate ``discretization-invariance," datasets such as FlowBench \cite{tali2024flowbench} (\cref{fig:flowbench}), the Well \cite{ohana2024well} and CFDBench \cite{luo2023cfdbench} provide simulations on hierarchical structures or varying mesh densities. These benchmarks explicitly test an operator's ability to generalize to unseen resolutions, a critical requirement for deploying operators in multi-scale engineering workflows. As shown in \cref{fig:amr}, these are used to model adaptive mesh problems.

\section{Applied problems: from dynamic forecasting to inverse design}
\label{sec:app}
While standardized benchmarks provide a rigorous testing ground for architectural validity, real-world deployment requires scaling these methods to systems with significantly higher complexity. To highlight the practical scope of neural operators, we categorize their applications by the nature of the problem being solved. Broadly, these fall into three categories: (1) Forecasting Dynamic Systems, where the goal is stability and long-horizon accuracy; (2) Interaction Modeling in Heterogeneous Media, where the goal is handling complex geometries and material properties; and (3) Inverse Problems and Design, where the operator acts as a differentiable surrogate for parameter estimation.

\subsection{Forecasting and time-evolution of dynamic systems} This category encompasses problems governed by time-dependent PDEs (e.g., Navier-Stokes, Atmospheric equations). The primary challenge here is mitigating error accumulation over long temporal rollouts.

\paragraph{Planetary and urban climate stability} The primary challenge in climate modeling is maintaining physical consistency over long autoregressive rollouts. At the planetary scale, FourCastNet \cite{pathak2022fourcastnet} achieves a massive 45,000$\times$ speedup over traditional Numerical Weather Prediction (NWP) systems for ensemble forecasting. However, it suffers from significant drift and energy dissipation when integrated over long time horizons, often requiring specialized multi-step loss functions to prevent unphysical blurring.

\begin{figure}[th!]
    \centering
    \includegraphics[width=0.8\linewidth]{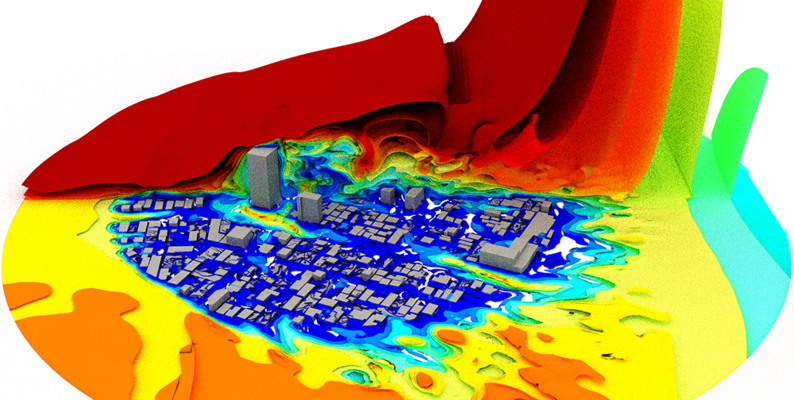}
    \caption{An illustration of the scale of the dataset in the CityFFD dataset, as used in \cite{peng2024fourier}}
    \label{fig:cityffd}
\end{figure}

At the urban scale, \cite{peng2024fourier} utilized FNOs to model micro-climates (wind, temperature) by training on CityFFD data (\cref{fig:cityffd}) generated via a semi-Lagrangian approach with the Smagorinsky Large Eddy Simulation (LES) model. While the operator achieved a 25$\times$ speedup with a relative error rate of only 5\% over 1200 time steps, it struggles to generalize to unseen city topologies without extensive retraining. This highlights the limitation of spectral methods in handling the sharp geometric discontinuities typical of urban canopies.

\paragraph{Coastal dynamics and multivariate complexity} Bridging fluid dynamics and disaster prevention, \cite{jiang2021digital} developed operator-based surrogates for the NEMO ocean model to predict sea surface height. By extending the FNO to learn multivariate dynamics (coupling wind, pressure, and water levels), they achieved a 45$\times$ speedup over the numerical solver, enabling real-time flood prediction. However, a persistent bottleneck in such ``Digital Twin" applications is data assimilation, effectively updating the operator's latent state with sparse, noisy real-time sensor data during an unfolding event without breaking the physical consistency of the predicted flow \cite{singh2024learning}.

\begin{figure}
\centering
\includegraphics[width=1\linewidth]{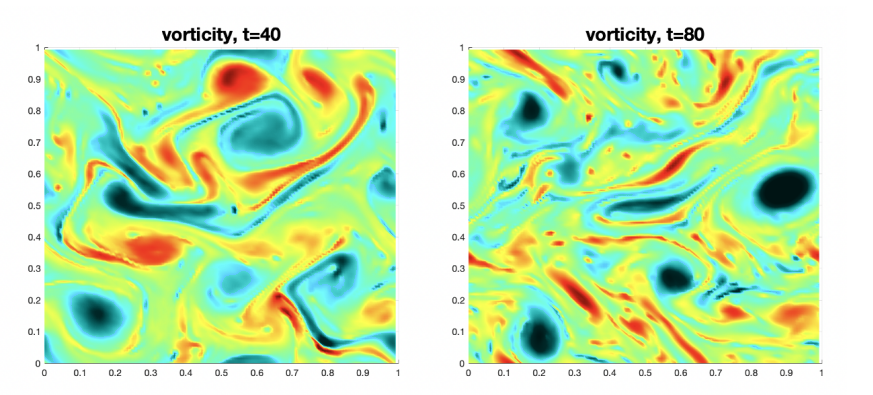}%
  \label{fig:chaosflow}%
\caption{This figure shows the simulation of Kolmogorov flow by the Markov neural operator (MNO) with an initial condition generated from a random Gaussian field. The model captures the energy spectrum that converges to the cascade rate of $k^{-5/3}$. Image reproduced from \cite{li2022fourier}}
\label{fig:apps}
\end{figure}

\paragraph{Turbulence and the spectral bias} For chaotic flows, a persistent bottleneck is the spectral bias of neural networks. \cite{renn2023forecasting} applied FNOs to forecast vortex shedding in cylinder wakes. While the model correctly captured large-scale shedding frequencies, it smoothed out fine-scale turbulent eddies, preventing the correct resolution of the energy cascade in high-Reynolds regimes. Even specialized architectures like the Markov Neural Operator \cite{li2021markov}, which successfully captures long-term attractor statistics for Kolmogorov flows (\cref{fig:apps}), are limited to ergodic systems and struggle to extrapolate to flow regimes outside the training distribution. 

\subsection{Modeling in heterogeneous and complex media} In this category of problems, the complexity stems from spatial heterogeneity. Key challenges involve learning to approximate sharp discontinuities found in multi-material interfaces or shocks. We discuss the following areas that focus on this class of problems. 

\paragraph{High-contrast porous media} Modeling multiphase flow in subsurface reservoirs involves permeability fields with sharp, high-contrast jumps. \cite{wen2022u} and \cite{diab2024u} successfully applied U-FNO and DeepONet architectures to coupled $CO_2$-water flow. The primary contribution of these works is the operator's ability to map static, irregular fields (permeability and porosity) directly to dynamic saturation profiles, effectively bypassing the iterative solvers required for Darcy's law.  However, these spectral-based models often exhibit ringing artifacts (oscillations) near sharp material interfaces due to the global nature of Fourier basis functions \cite{cavallazzi2025walsh}. Furthermore, generating high-fidelity multiphase training data remains computationally prohibitive, often leading to overfitting on limited geological realizations\cite{jiang2023use}.

\paragraph{Inelastic deformation and impact} In solid mechanics, \cite{liu2022learning} utilized operators to model the inelastic deformation of polycrystalline solids (e.g., magnesium) under high-velocity impact. By learning the kinetic relation between crystallographic orientation and stress, the model bypasses expensive crystal-plasticity FEM simulations. While effective for homogenization, a major bottleneck remains: resolving the localized stress concentrations at grain boundaries requires immense spectral resolution. This ``scale mismatch", where the grain boundary is orders of magnitude smaller than the impact zone, can negate the computational advantage of the operator compared to adaptive FEM \cite{khorrami2024divergence}.
\begin{figure}
    \centering
    \subfloat[The figure represents the degree of cure predicted by the FEM method vs. the same done by FNO.]{%
        \includegraphics[width=0.4\textwidth]{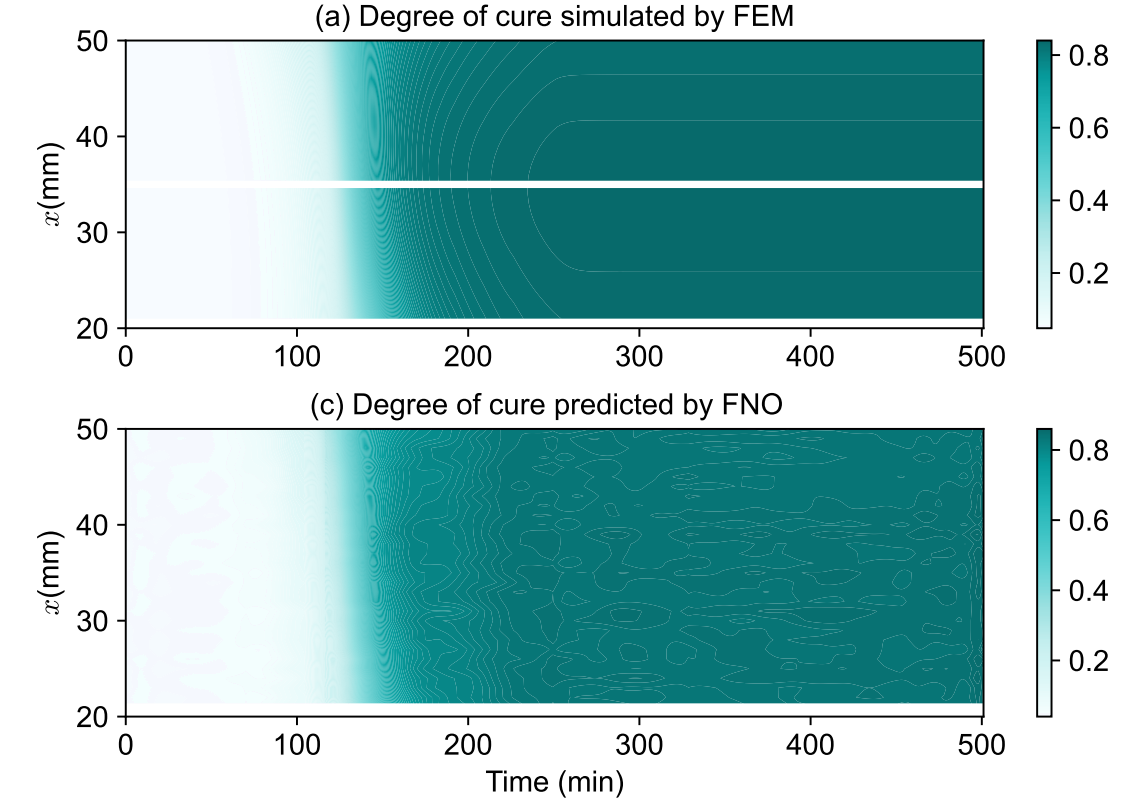}%
        \label{fig:curefig}%
    }
    \hfil 
    \subfloat[These graphs represent the degree of cure predicted by FNO and FEM methods at x = 35mm and 21mm.]{%
        \includegraphics[width=0.4\textwidth]{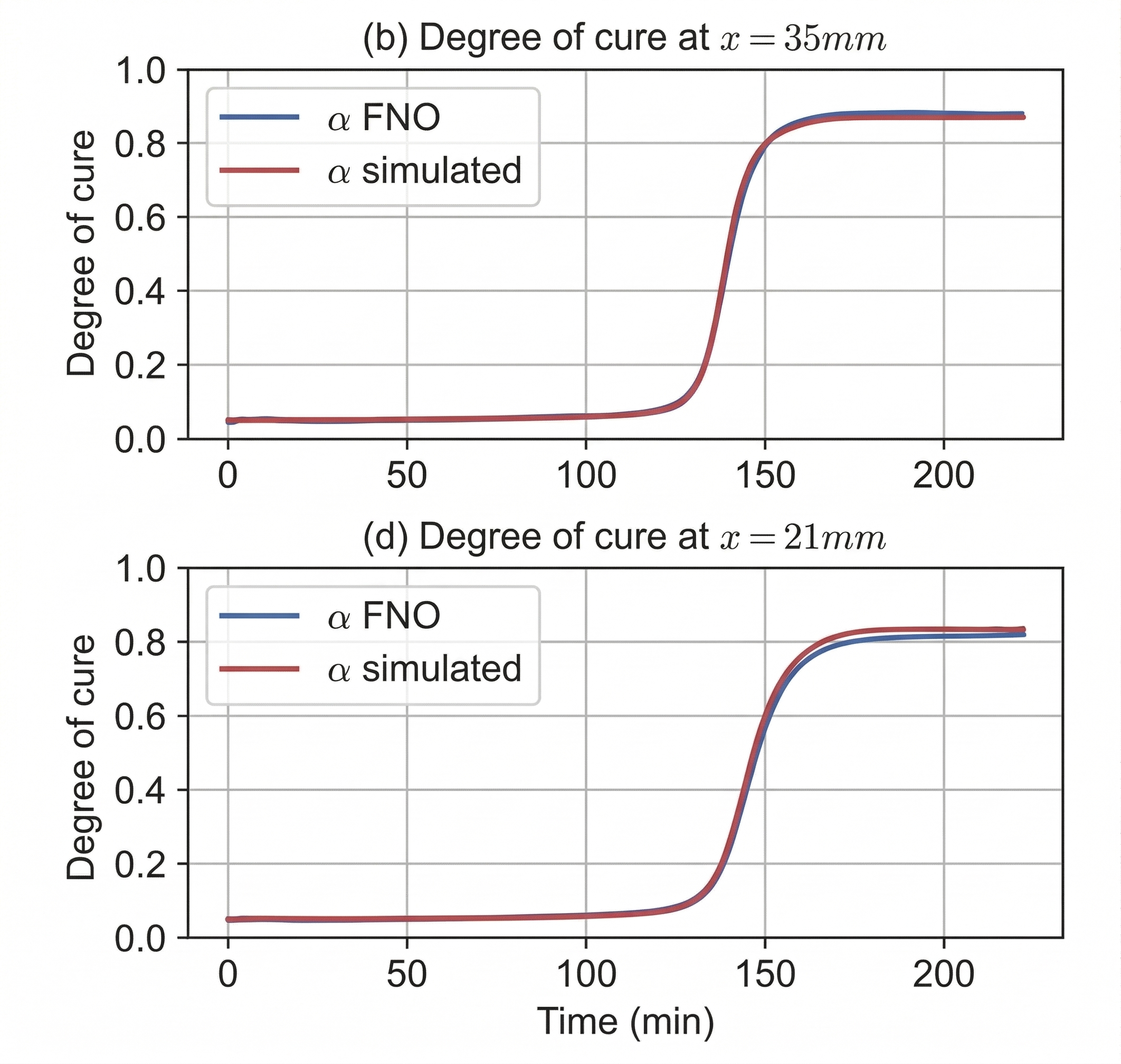}%
        \label{fig:cureGraph}%
    }
    
    \caption{Performance visualizations on thermochemical curing applications, as shown in \cite{chen2021residual}}
    \label{fig:thermo_apps}
\end{figure}

\paragraph{Thermochemical curing} Manufacturing composites requires precise thermal control to prevent defects. \cite{chen2021residual} applied residual-FNOs to the heat transfer equations governing this process. This is illustrated in \cref{fig:thermo_apps}. The operator successfully captures the non-linear relationship between the curing cycle history and the internal temperature gradient, allowing for rapid optimization of the heating process. However, accurate prediction depends heavily on the assumption of homogenized material properties. A persistent challenge is extending these models to detect or predict local defects (e.g., delamination or voids) where the continuum assumption breaks down and the thermal conductivity becomes discontinuous.

\subsection{Inverse problems, design, and optimization} The third class leverages the differentiability of operators. Unlike discrete solvers, operators can be differentiated via backpropagation, making them ideal for inverse problems (finding inputs from outputs) or design optimization. However, the utility of these methods is constrained by the ill-posedness of inverse problems and the theoretical difficulty of enforcing constraints on high-dimensional manifolds.

\paragraph{Seismic inversion} \cite{yang2021seismic} demonstrated that differentiating through an FNO enables Full Waveform Inversion (FWI) without an adjoint solver, recovering velocity maps directly from seismic data. The method effectively learns a "prior" over possible geologies, accelerating convergence by orders of magnitude. However, this advantage is also a bottleneck: if the subsurface structure deviates from the learned distribution (out-of-distribution), the inversion often converges to spurious artifacts \cite{ma2025effective}. Unlike classical physics-based inversion which relies solely on the wave equation, operator-based inversion is biased by the training dataset's representation of geology.

\paragraph{Physics-violation in bio-design} In bio-molecular design, \cite{omar2023protein} and \cite{van2024pmipred} employed surrogates to minimize protein binding energy, effectively solving an inverse design problem. While the operator vastly accelerates the search for optimal configurations, a key risk is physical validity. The operator, being an approximation, may smooth over high-energy ``clashes" (steric hindrances) or Van der Waals singularities that a rigid physics engine would reject. Consequently, the ``optimal" designs found by differentiating the operator may be adversarial examples that exploit the network's approximation errors rather than true energy minima \cite{schwalbe2021differentiable}.

\begin{figure}[!ht] 
    \centering
    
    \begin{minipage}{0.48\linewidth}
        \centering
        \includegraphics[width=\linewidth]{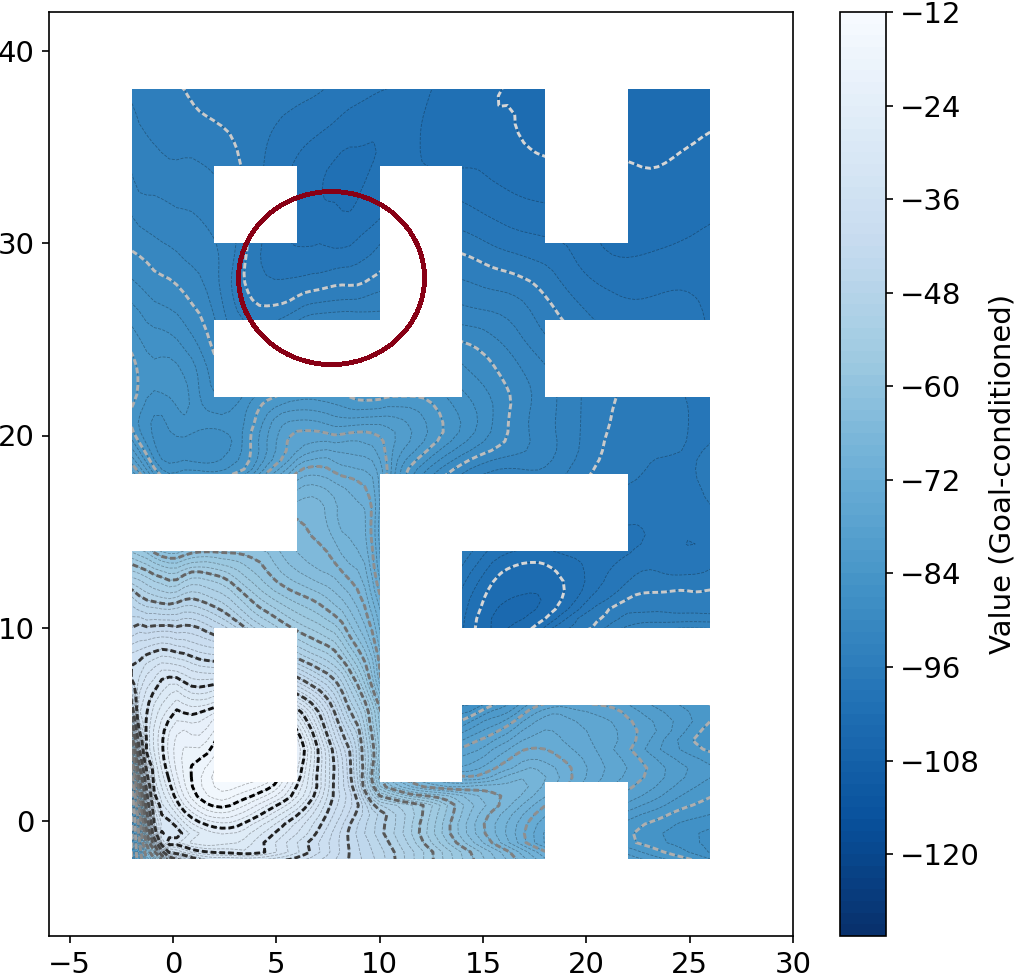}
        \vspace{2pt} 
        \centerline{(a) GCIVL}
    \end{minipage}
    \hfill
    \begin{minipage}{0.48\linewidth}
        \centering
        \includegraphics[width=\linewidth]{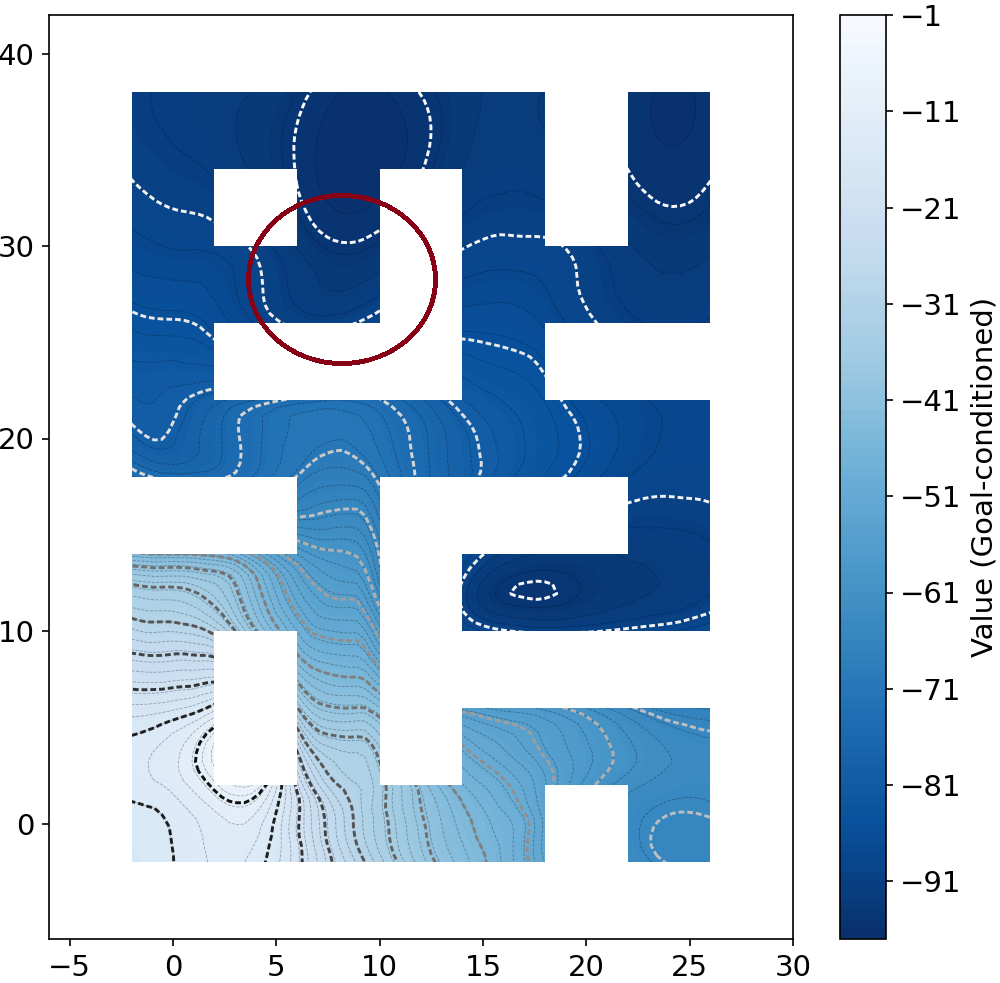}
        \vspace{2pt} 
        \centerline{(b) GCIVL+PINN}
    \end{minipage}

   \caption{%
        Reproduced from \cite{viswanath2026physics}, these plots showcase the impact of physics-informed learning on value landscape in reinforcement learning tasks. The plot represents the value function over a maze like environment, with PINN constraints improving the contours of the value function}
    \label{fig:horizontal_ablation}
\end{figure}

\paragraph{Physics-informed robotics and planning} Physics is an inherent component of motion planning, where navigation is often framed as finding geodesics or optimal control policies. Recent works \cite{bhaskara2024trajectory, peng2023graph} have adapted neural operators for multi-agent path planning and collaboration. However, integrating rigorous physics constraints, such as Eikonal equations or the Hamilton-Jacobi-Bellman (HJB) models for planning and reinforcement learning, remains difficult. These formulations often struggle with gradient discontinuities arising from obstacle boundaries and suffer from stability issues when solving the Eikonal equation on irregular domains \cite{chen2025manifold, giammarino2025physics,ni2025physics,ni2022ntfields}. Recent works, such as \cite{giammarino2025physics, viswanath2026physics} have shown that Eikonal and HJB based constraints can improve the value function manifold in reinforcement learning tasks, as seen in \cref{fig:horizontal_ablation}. Moreover, extending these methods from low-dimensional navigation to high-dimensional configuration spaces, such as 7DOF Lie groups for robotic manipulators, remains a non-trivial open problem \cite{ren2025physics}.

\begin{figure}[th!]
    \centering
    \includegraphics[width=0.75\linewidth]{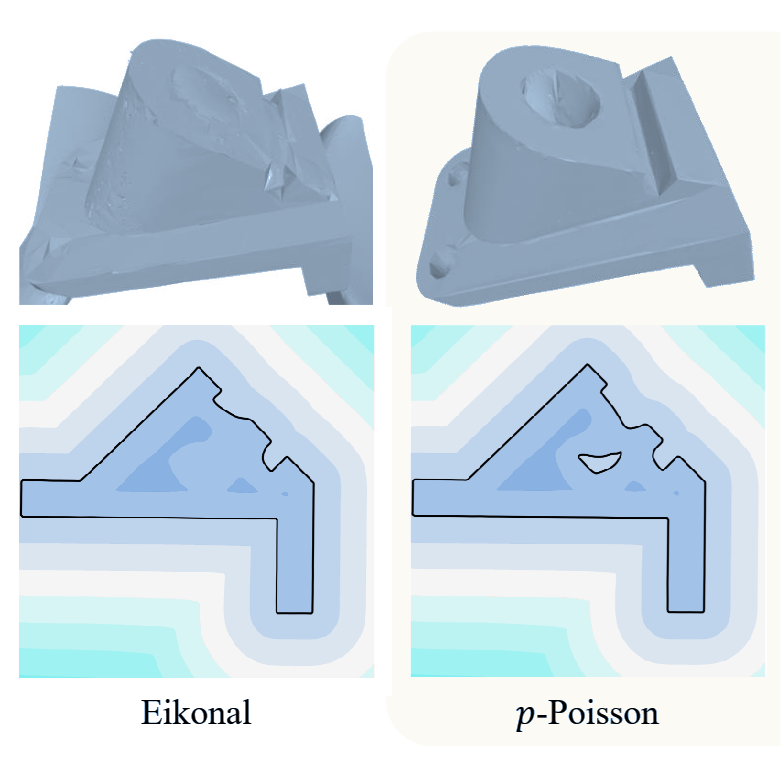}
    \caption{Illustration of level set contours from different physics informed strategies, as presented in \cite{park2023p}}
    \label{fig:surfacerecon}
\end{figure}

\paragraph{Physics-informed surface reconstruction} Poisson equations and Eikonal constraints are often used in surface reconstruction. To enable resolution-agnostic modeling of the zero-level sets that describe a surface, recent approaches combine data driven neural field operators with physics informed losses, such as Eikonal or p-Poisson \cite{park2023p} or directly train neural operators to approximate the function \cite{andrade2023neural} (\cref{fig:surfacerecon}).

\begin{table*}[h]

\resizebox{1\textwidth}{!}{
\begin{tabular}{llll}
\hline
\textit{Model}                                         & \textit{Advantages}                                                                                                                                                                                                                                                                                                                                     & \textit{Limitations}                                                                                                                                                                                                                                                                                           & \textit{Applications}                                                                                                                                                                                                          \\ \hline
\multicolumn{1}{|l|}{\textbf{Fourier neural operator}} & \multicolumn{1}{l|}{\begin{tabular}[c]{@{}l@{}}Faster\\ Resolution invariant\\ discretization invariant\\ Data driven\\ Doesn't need to know the underlying PDE\\ Zero shot Super resolution \cite{li2020fourier} \end{tabular}}                                                                                                                                              & \multicolumn{1}{l|}{\begin{tabular}[c]{@{}l@{}}Parameterization may lead to opaque outputs \\ and aliasing errors \cite{fanaskov2022spectral} \\ Only works on rectangular domains with uniform \\ meshes \cite{li2022fourier}\\ Overfits with deeper networks\\ Susceptible to Vanishing Gradient \cite{rahman2022generative}\\ Constrained by availability of training data. \cite{li2021physics}\end{tabular}} & \multicolumn{1}{l|}{\begin{tabular}[c]{@{}l@{}}Burger's Equation\\ Darcy Flow Equation\\ Navier Stokes Equation \cite{li2020fourier} \\ Coastal Flood modelling \cite{jiang2021digital}\\ Photoacoustic Equation \cite{guan2021fourier}\\ Chaotic systems \cite{li2021markov} \\ Seismic wave progressions \cite{yang2021seismic} \end{tabular}} \\ \hline
\multicolumn{1}{|l|}{\textbf{GANO/UNO}}                & \multicolumn{1}{l|}{\begin{tabular}[c]{@{}l@{}}Memory efficient implementations of\\ deeper networks\\ Optimized for Polish and Banach spaces \\ Works well for bounded norms in infinite spaces\\ Good at learning probability measures\\ Don't suffer from modal collapse \cite{rahman2022generative} \end{tabular}}                                                                & \multicolumn{1}{l|}{\begin{tabular}[c]{@{}l@{}}Only works on rectangular domains with uniform\\ meshes \cite{li2022fourier} \\ \\ Parameterization may lead to opaque outputs and \\ aliasing errors. \cite{fanaskov2022spectral}\end{tabular}}                                                                                                                  & \multicolumn{1}{l|}{\begin{tabular}[c]{@{}l@{}}Volcanic deformations \cite{rahman2022generative} \\ Video Interpolation \cite{viswanath2022nio}\\ Dyadic Human motion prediction \cite{rahman2022pacmo}\end{tabular}}                                                                                      \\ \hline
\multicolumn{1}{|l|}{\textbf{DeepONet}}                & \multicolumn{1}{l|}{\begin{tabular}[c]{@{}l@{}}Accurately approximate mappings\\ between infinite dimensional banach\\ spaces\\ Learns oscillatory continuous functions\\ Can learn the mapping from high frequency\\  functions to low frequency functions \cite{lu2019deeponet}\end{tabular}}                                                                             & \multicolumn{1}{l|}{\begin{tabular}[c]{@{}l@{}}Requires high amount of training data\\ May fail to learn underlying physical\\ principles\\Not resolution invariant: Only takes input function at a \\fixed discretization\\ \end{tabular}}                                                                                                                                                        & \multicolumn{1}{l|}{\begin{tabular}[c]{@{}l@{}}Material Physics \cite{goswami2022physics} \\ Electrodynamics \cite{cai2021deepm} \\ Aerothermodynamics \cite{sharma2021application} \\ Medical imaging \cite{desanctis1987aortic} \\ Effects of seismic waves on buildings \cite{liu2021multiscale}\end{tabular}}                                                 \\ \hline
\multicolumn{1}{|l|}{\textbf{Graph neural operator}}   & \multicolumn{1}{l|}{\begin{tabular}[c]{@{}l@{}}Learns long range dependencies in graph like data\\ Linear time complexity\\ Discretization invariant\\ Can learn Mesh invariant solutions \cite{li2020multipole} \end{tabular}}                                                                                                                                                 & \multicolumn{1}{l|}{\begin{tabular}[c]{@{}l@{}}Outperformed by Fourier neural operator on all PDEs \\ with regular meshes. \cite{li2020fourier} \end{tabular}}                                                                                                                                                                       & \multicolumn{1}{l|}{\begin{tabular}[c]{@{}l@{}}Burgers Equation\\ Darcy Flow Equation \cite{li2020multipole}\\ Protein dynamics in SARS-COV-2 \\ virus \cite{trifan2021intelligent}\end{tabular}}                                                                                  \\ \hline
\multicolumn{1}{|l|}{\textbf{PINO}}                    & \multicolumn{1}{l|}{\begin{tabular}[c]{@{}l@{}}Doesn't suffer from generalization \\ errors that other operators suffer from\\ Overcomes the limitations of purely physics based\\ and purely data driven approaches. \\ Incorporate constraints at different resolutions - \\ combine coarse resolution data and high resolution\\ data. \cite{li2021physics}\end{tabular}} & \multicolumn{1}{l|}{\begin{tabular}[c]{@{}l@{}}Has not been tested rigorously on High dimensional \\ PDEs \cite{li2021physics} \end{tabular}}                                                                                                                                                                                        & \multicolumn{1}{l|}{\begin{tabular}[c]{@{}l@{}}Long Temporal Transient Flow\\ Kolmogorov Flows\\ Wave Equation\\ Non-Linear Shallow Water Equation \cite{li2021physics}\end{tabular}}                                                               \\ \hline
\multicolumn{1}{|l|}{\textbf{Adaptive FNO}}            & \multicolumn{1}{l|}{\begin{tabular}[c]{@{}l@{}}Powerful generative model \cite{pathak2022fourcastnet} \\ Efficient Token Mixer \\Adaptive Weight sharing among tokens \\Quasi-Linear Time Complexity \\ highly parallelized \\ Outperforms self-attention mechanisms  \cite{guibas2021adaptive} \end{tabular}}                                                                                                                                                                                                                                                                                                          & \multicolumn{1}{l|}{\begin{tabular}[c]{@{}l@{}} Can be modified through wavelet transforms \\
to better capture locality \cite{guibas2021adaptive} \end{tabular}}                                                                                                                                                                                                   & \multicolumn{1}{l|}{\begin{tabular}[c]{@{}l@{}}Climate Modelling\\ Weather forecast\\ Hurricane prediction \cite{pathak2022fourcastnet}\\ Generative imaging \cite{guibas2021adaptive}\end{tabular}}                                                                                  \\ \hline
\multicolumn{1}{|l|}{\textbf{geo-FNO}}                 & \multicolumn{1}{l|}{\begin{tabular}[c]{@{}l@{}}geometry Aware\\ Input can be irregular meshes, point clouds\\ As fast as FNO but more efficient and accurate \cite{li2022fourier} \end{tabular}}                                                                                                                                                                              & \multicolumn{1}{l|}{\begin{tabular}[c]{@{}l@{}}Only been tested on regular homeomorphic topologies\\ Can be potentially expanded into PINOs but hasn't been\\ empirically verified. \cite{li2022fourier} \end{tabular}}                                                                                                              & \multicolumn{1}{l|}{\begin{tabular}[c]{@{}l@{}}Structural and Fluid Mechanics\\ Problems\\ Euler's equation for Airfoil flow \cite{li2022fourier}\end{tabular}}                                                                                     \\ \hline
\multicolumn{1}{|l|}{\textbf{Implicit FNO}}            & \multicolumn{1}{l|}{\begin{tabular}[c]{@{}l@{}}Doesn't suffer from Vanishing Gradient\\ Less prone to overfitting\\ Hidden Layer parameters are independent\\ Has the ability to learn material responses directly\\ from DIC displacement tracking measurements. \cite{you2022learning} \end{tabular}}                                                                         & \multicolumn{1}{l|}{\begin{tabular}[c]{@{}l@{}}Has long training times despite having fewer parameters\\ due to the iterative algorithm used for learning. \cite{you2022learning} \end{tabular}}                                                                                                                                       & \multicolumn{1}{l|}{\begin{tabular}[c]{@{}l@{}}Model Heterogenity and\\ Material Defects in anisotropic\\ and hyperelastic setting\\ Porous Medium Flow\\ Fracture mechanisms \cite{you2022learning} \end{tabular}}                                    \\ \hline
\multicolumn{1}{|l|}{\textbf{Multiwavelet FNO}}        & \multicolumn{1}{l|}{\begin{tabular}[c]{@{}l@{}}Compact Representation of data\\ Resolution-independent solutions\\ Learn complex dependencies \cite{gupta2021multiwavelet} \end{tabular}}                                                                                                                                                                                             & \multicolumn{1}{l|}{\begin{tabular}[c]{@{}l@{}}Performance degrades if the Kernel\\ used for data generation is changed. \\ Cannot generalize to high frequency \\ signals from low frequency ones \cite{gupta2021multiwavelet} \end{tabular}}                                                                                               & \multicolumn{1}{l|}{\begin{tabular}[c]{@{}l@{}}Burgers Equation (1D)\\ Navier-Stokes Equation (2D)\\ Darcy Flow Equation (2D)\\ Korteweg-de Vries Equation (1D) \cite{gupta2021multiwavelet} \end{tabular}}                                                  \\ \hline
\multicolumn{1}{|l|}{\textbf{Spectral FNO}}            & \multicolumn{1}{l|}{\begin{tabular}[c]{@{}l@{}}Doesn't suffer from aliasing errors\\ Lossless operations on Functions\\ Preserves the structure of the functions \cite{fanaskov2022spectral} \end{tabular}}                                                                                                                                                                          & \multicolumn{1}{l|}{\begin{tabular}[c]{@{}l@{}}Doesn't perform well on Burger's Equations\\ Suffers from Gibbs Phenomenon\\ Only works on smooth input/output\\ Basis functions used are non-adaptive \cite{fanaskov2022spectral} \end{tabular}}                                                                                            & \multicolumn{1}{l|}{\begin{tabular}[c]{@{}l@{}}Basic Integration, Differentiation\\ Parametric ODEs\\ Elliptic Equations\\ KdV Equation\\ Non-Linear Schrodinger Equation \\
\cite{fanaskov2022spectral}\end{tabular}}                                        \\ \hline
                                                       &                                                                                                                                                                                                                                                                                                                                                         &                                                                                                                                                                                                                                                                                                                &                                                                                                                                                                                                                                \\ \hline
\end{tabular}
}
\caption{The table highlights the key advantages and limitations of the classical operator-based neural architectures for solving PDE and other physics problems}
\label{tab:Summary}
\end{table*}

\section{Scalability and Generalizability}

\label{scalability}
The key advantages of data-driven techniques are scalability and generalizability. Finite element approaches are generally slow and imprecise as they suffer from a trade-off between mesh resolution and computation time. The higher the resolution, the higher the computation cost per evaluation. Moreover, the results of a computation are only valid for a single instance of a PDE. Changing initial or boundary conditions or any other parameter requires the expensive computation to be re-run.  

Neural network-based methods are capable of approximating a PDE with decent accuracy. Although training them is computationally expensive, each computation of the forward evaluation is much faster than the conventional approaches. Some of these methods, such as the ConvPDE-UQ framework, are even able to parameterize the initial and boundary conditions as part of the input to the model in order to enable computation of any system within a given family of PDEs on any domain without having to retrain the model (\cite{winovich2019convpde}). However, neural network approaches are still mesh-dependent. The resolution that can be achieved from the model is determined by the resolution of the training data, which is computed via conventional methods. Thus, the computational cost for training is still sensitive to resolution. This is where a neural operator is advantageous, as it is capable of learning a mapping between infinite-dimensional spaces. The result is that they are resolution invariant in that they can be trained with data that has a relatively low resolution and will still be able to evaluate at a higher resolution with the same error rate as in low resolution. \cite{li2021physics} demonstrated that their neural operator variant, the Fourier neural operator, is capable of accurately learning PDEs with zero-shot super-resolution. They further proved that the Fourier neural operator achieves superior accuracy compared to neural network-based solvers while still enjoying the same benefits in terms of fast computation of the forward evaluation and generalization to any instance within a PDE family.

All of these features of FNOs- low evaluation cost, zero-shot super resolution, and generalization- have significant implications in terms of computational efficiency. \cite{li2021physics} highlight this by presenting the Bayesian Inversion Problem, where they use a function space Markov chain Monte Carlo (MCMC) method (\cite{cotter2013article}) to draw samples from the posterior distribution of the initial vorticity in Navier-Stokes. MCMC is a set of algorithms for sampling data from distributions. The approach involves constructing Markov chains for the desired distribution, and a sample of this distribution would be a set of states of the Markov chain. Metropolis-Hastings is a well-known MCMC algorithm. In this experiment, they compare FNOs and conventional solvers. While both are able to achieve similar results, they vary substantially in terms of computation time. The MCMC using the traditional solver took 2.2 seconds per computation, whereas the FNO took only $0.005$s per computation. The FNO, however, requires a one-time training, which in this scenario took approximately 12 hours. The traditional solver, on the other hand, requires no such training. Hence, for a small number of data points, the conventional solver is more efficient. The benefits of FNOs are instead realized at a larger scale because the model only has to be trained once and can be used to quickly evaluate any instance of the PDE. To illustrate this point, let T represent the total computation time in seconds to make n evaluations. For the traditional solver, the computation time has the form of a fixed rate  (\(T=2.2n\) in this particular setting), whereas, for FNO, computation time would behave as a lower rate plus a one-time cost (\(T=43200+0.005n\) in their setup). The authors generated 30,000 data points using each method. Using the FNO, this took 12 hours for training plus an additional 2.5 minutes for the 30,000 forward evaluations, whereas the traditional solver took a total of 18 hours for the same number of evaluations (\cite{li2021physics}).

These advantages in computation time extend to cost savings, especially at a large scale. To illustrate the difference in cost at a large scale, consider the cost of running both solvers with the same experimental setup as above on a p3.2xlarge EC2 instance on AWS, which has 8 vCPU and 61 GiB (or approx. 65 GB) of memory. The P3 instances feature the NVIDIA V100 GPU, making it the most similar of the AWS compute options to the NVIDIA V100 GPU with 16 GB memory used by \cite{li2021physics}. For the purposes of this demonstration, assume these instances have the same hardware performance as the hardware used by Li et al. The on-demand rate for this instance, which is the cheapest of the P3s, is \$3.06 per hour. Thus, training the FNO would cost approximately \$36.72. Once trained, the FNO can theoretically perform up to 720000 computations per hour, while the conventional solver would take 440 hours to do the same number of evaluations. For 100,000 evaluations, the FNO would cost under \$40, including training time, whereas the conventional solver would cost approximately \$187. Beyond the cost savings, this computational efficiency has important implications for environmental impact, as cloud computing is known to consume significant energy, resulting in a large carbon footprint.  

In designing mechanical systems involving aerodynamics or fluid dynamics, it is common for engineers to have to compute the forward operation of Navier-Stokes several thousand times, varying the coefficients with each evaluation in order to recover certain properties of the system. \cite{Fourestey2005article}, for example, investigated inverse problems of the Navier-Stokes equations in the context of designing bridge decks under wind loads. Solving the inverse problem of a given PDE is also critical to tuning predictive modeling systems such as those for weather and climate modeling (\cite{Kashinath2021article}). Using conventional numerical methods to solve inverse problems of a given PDE can be prohibitively slow and expensive. The scalability of FNOs can make this problem not only feasible but cost-effective, which has significant implications for mechanical design.  

\section{Future work - Adaptability}
In this paper, we have illustrated the advantages that data-driven approaches have over traditional numerical solvers in terms of computational performance, as well as their superior accuracy in certain settings. The remaining question to be answered is how these benefits can be realized in academic research and commercial applications. The development of numerical methods such as FDM and FEM began as early as the 1940s \cite{hrennikoff1941solution}. However, these methods did not see widespread use until the 1960s when open-source programs for FEM began to be developed, such as Nastran developed by NASA \cite{butler1971nastran} and SAP IV developed at UC Berkeley \cite{gran1978nastran}. Since then, many tools for the numerical modeling of physical systems have been built and made widely available to researchers and engineers, including MATLAB, COMSOL, and Autodesk Simulation. The question now becomes- \textit{how can the same be done with FNO-based solvers?}  

As FNOs are still relatively new computational methods, there is more work that is needed to develop them further, and more variants of them are likely to be developed over the next few years. Nevertheless, existing variants already have the potential to greatly accelerate physics-based modeling. However, as with many newly developed neural networks, they are unlikely to see practical use or widespread adoption by physics researchers until they are integrated into a software package that can reliably be used without a firm background in machine learning. Developing this software and enabling it to be used for practical applications would further motivate future research and development of these approaches for PDE approximation. An important factor in achieving this is open-source. Li et al. have already made their code open-source for their work in \cite{li2020fourier}. Open-source accelerates development and enables standardization in software packages. This standardization is central to the repeatability of results and is, hence, critical for research.  

It is important to note that numerical solvers are still required to produce the training data for FNOs. As with all machine learning approaches, the quality of this data, in terms of the size and spread of training data, as well as accuracy, impacts the performance of the FNO. In this respect, machine learning paradigms such as that of active learning \cite{settles2012active}, which uses learning theoretic arguments to come up with strategies to select training data, can be used to improve the sample complexity (a term used in machine learning for the number of samples required to achieve a particular accuracy) of learning based techniques. In particular, the learning algorithm would decide on the pairs of $\{u_i\}_{i=1}^N$ for which the solver should generate the corresponding $\{f_i\}$ or $\{a_i\}$ with the aim of selecting the right pairs such that the network is able to approximate the solver function with the lowest number of pairs $N$. Meta-learning is another paradigm that can be used to improve the sample complexity of learning algorithms. This method aims to learn some underlying connections to different machine-learning tasks. While there are several notions of meta-learning, a popular algorithm is MAML \cite{finn2017model}, which aims to learn an initialization for gradient-based learning approaches from different tasks sharing a common structure. This learned initialization, when used for the gradient-based learning approaches, allows for better sample complexity \cite{saunshi2020sample}. 

Another paradigm is zero-shot and few-shot generalization and in-context learning, which has risen in prominance due to the popularity of large-language models and foundational models \cite{yang2023icon, yang2025iconlm, yang2024pdeicon, cao2026vicon, wu2026gicon, zhang2025probabilistic, meng2026execution, cole2026probability}. This new paradigm focuses on efficient ways to generalize during inference to avoid expensive re-training. 

Due to the dependency of FNOs on numerical solvers, one potential future for FNOs would be to integrate them into FEM software packages. This would make them more accessible to physics researchers, allowing them to leverage the computational benefits of FNOs to accelerate their work. The software could train an FNO for a user-inputted PDE with data it produces using FEM and a sampling method based on active learning or meta-learning. The trained FNO would then be used in the computation of the forward evaluation with a mesh and other parameters defined by the user via the same interfaces they use today, as the software would handle the parameterization of those inputs into the structure expected by the FNO. Of course, there are many variants of FNOs, each created for different types of systems involving PDEs, where they achieve superior performance over the vanilla version. Users of FEM software would not be aware of the distinctions between them. Hence, this software would have to be able to provide the user with some recommendations of which variant, if any, to use in which scenarios. A deeper study of the performance of different variants across different use cases would be required in order to build this knowledge base. This paper is a starting point for that. 

While the accuracy of FNO-based solvers is invariant to resolution, it is still true that a given FNO-based solver could have higher accuracy on certain systems over others due to the non-deterministic nature of machine learning. This would be an additional consideration for researchers using this software. One approach to contend with this would be to have the software also produce test data along with the training data and provide a report of accuracy on this test set so that the researcher can weigh the accuracy-efficiency trade-off between FNO and conventional methods in order to decide which approach to use. If they opt for the FNO approach, they could also include this report in their findings. 

\section{Conclusion}
The goal of this paper was to characterize the numerous methods that have been developed for approximating PDEs in order to highlight the potential opportunities that exist for further development of these technologies. We have provided a dense overview of these methods, from conventional solvers to more novel approaches developed in recent years that leverage neural networks to approximate the mapping between finite spaces, as well as those capable of approximating the mapping between function spaces by leveraging the neural operator. We have provided details about several variants of each type of PDE solver and have compared the performance of these variants across several applications, particularly those in the field of physics. We have also articulated the advantages of data-driven solvers as compared to conventional solvers and the implications this could have for the future of physics modeling. Furthermore, we proposed a high-level vision for how this technology could eventually be used to accelerate computation-based physics research. To return to our central question: is data all you need? It is increasingly apparent that the future lies in synergy. Data-driven surrogates are poised to complement traditional solvers, working in tandem to overcome the critical bottlenecks of physics modeling.

\bibliographystyle{IEEEtran}
\bibliography{main}
\appendix
\newpage
 
\vspace{11pt}

\begin{IEEEbiography}[{\includegraphics[width=1in,height=1.25in,clip,keepaspectratio]{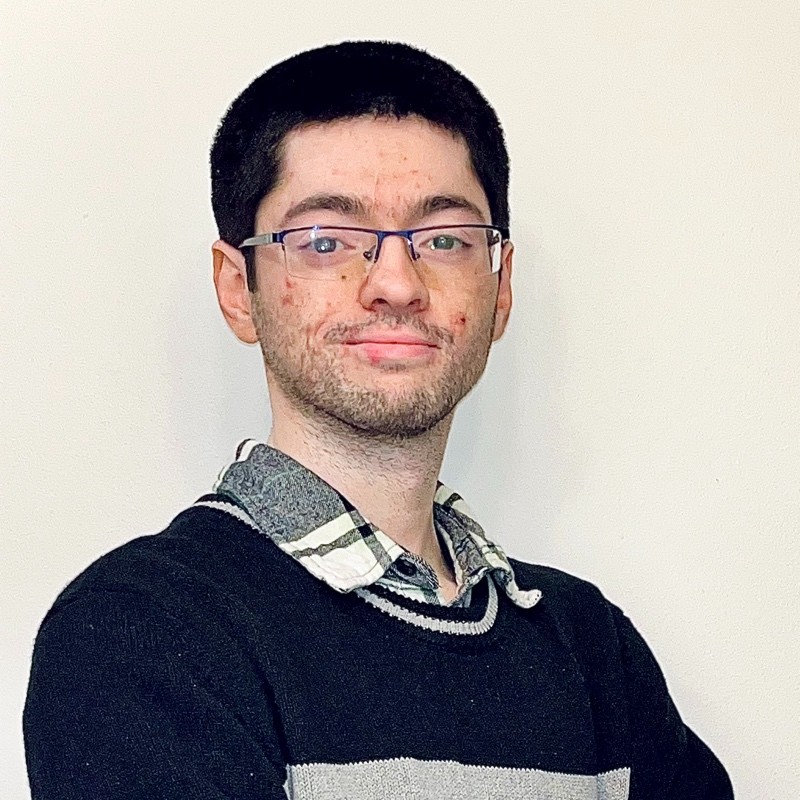}}]{Hrishikesh Viswanath}
is a PhD student at Purdue University specializing in stochastic methods in operator learning and Physics-Informed Neural Networks (PINNs) for modeling manifolds in graphics and robotics.
\end{IEEEbiography}

\begin{IEEEbiography}[{\includegraphics[width=1in,height=1.25in,clip,keepaspectratio]{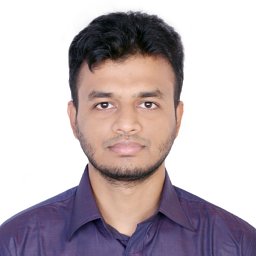}}]{Md Ashiqur Rahman}
is a PhD Student at Purdue University researching equivariance and invariance methods in operator learning problems.
\end{IEEEbiography}

\begin{IEEEbiography}[{\includegraphics[width=1in,height=1.25in,clip,keepaspectratio]{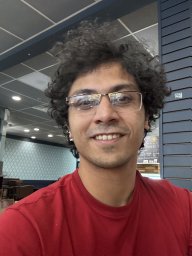}}]{Abhijeet Vyas}
is a PhD student at Purdue University working in min-max optimization. He is privileged to be guided by Prof. Brian Bullins.
\end{IEEEbiography}

\begin{IEEEbiography}[{\includegraphics[width=1in,height=1.25in,clip,keepaspectratio]{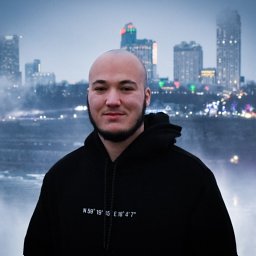}}]{Andrey Shor}
is a Machine Learning Engineer at Burns and McDonnell. He broadly works on data engineering pipelines and learning methods for NLP, with an interest in interpretability and physical modeling in reinforcement learning for robotics and NLP.
\end{IEEEbiography}

\begin{IEEEbiography}[{\includegraphics[width=1in,height=1.25in,clip,keepaspectratio]{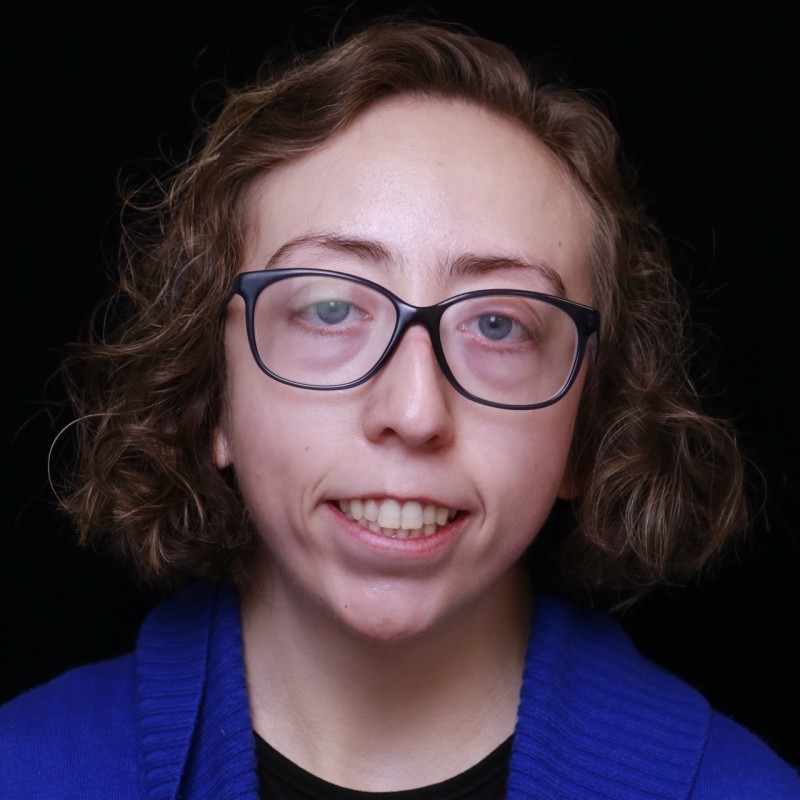}}]{Beatriz Medeiros}
is a PhD student at The University of Cambridge working at the intersection of artificial intelligence and physical modeling. In particular, her research is focused on developing neural operator architectures to solve parameterized partial differential equations on complex domains. Additionally, she has 5 years of experience working as a software development engineer at Amazon.
\end{IEEEbiography}

\begin{IEEEbiography}[{\includegraphics[width=1in,height=1.25in,clip,keepaspectratio]{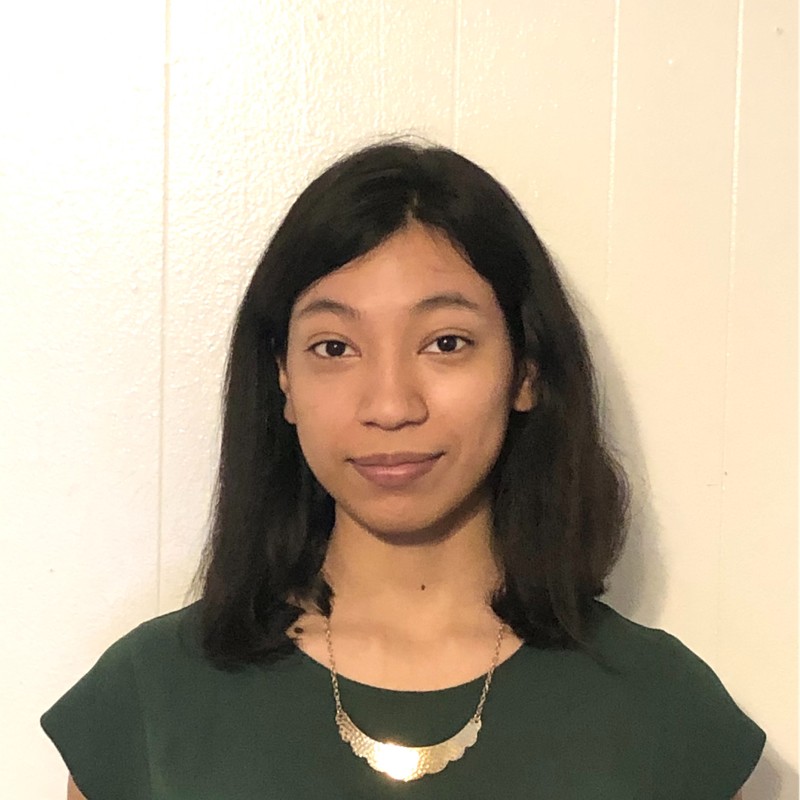}}]{Stephanie Hernandez}
received the B.S. degree in materials science from Johns Hopkins University. She is currently a Materials Engineer at Lockheed Martin.
\end{IEEEbiography}

\begin{IEEEbiography}[{\includegraphics[width=1in,height=1.25in,clip,keepaspectratio]{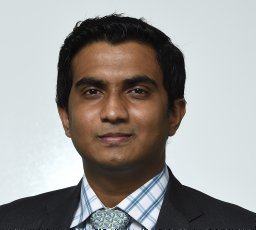}}]{Suhas Eswarappa Prameela}
is an Assistant Professor (tenure track) in the Department of Materials Science and Engineering at the John and Marcia Price College of Engineering, College of Mines and Earth Sciences. He previously held dual postdoctoral fellowships at the Massachusetts Institute of Technology (MIT) from 2022 to 2024, serving as an MIT Aeronautics and Astronautics Distinguished Postdoctoral Fellow in the Department of Aeronautics and Astronautics and as an MIT Engineering Excellence Postdoctoral Fellow in the Department of Materials Science and Engineering. Dr. Prameela earned his Ph.D. in Materials Science and Engineering from Johns Hopkins University (2016–2022) and his M.S. in the same field from Arizona State University (2014–2016).
\end{IEEEbiography}

\begin{IEEEbiography}[{\includegraphics[width=1in,height=1.25in,clip,keepaspectratio]{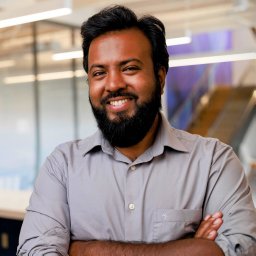}}]{Aniket Bera}
is an Associate Professor in the Department of Computer Science at Purdue University. He directs the interdisciplinary IDEAS Lab—Intelligent Design for Exploration and Augmented Systems—a robotics‐first group that studies how embodied agents plan, perceive, and collaborate safely with people. The lab’s emphasis is on motion planning with guarantees, human–robot collaboration, multi‐robot coordination, and manipulation in complex, dynamic environments. Computer vision powers perception and mapping (SLAM at scale, multi‐sensor fusion, and 3D scene representations), while graphics/VR supports simulation, evaluation, and human‐in‐the‐loop studies.
\end{IEEEbiography}

\vspace{11pt}

\vfill
\section{Appendix}
\begin{table}[h]
    \centering
    \resizebox{9cm}{!}{\begin{tabular}{c|c}
        \hline
        \textbf{Model} & \textbf{Link} \\
        \hline
        \textit{FNO} &  \href{https://github.com/neural-operator/fourier\_neural\_operator}{https://github.com/neural-operator/fourier\_neural\_operator}\\
        \textit{FourCastNet} & \href{https://github.com/NVlabs/FourCastNet}{https://github.com/NVlabs/FourCastNet}\\
        \textit{GANO} & \href{https://github.com/kazizzad/GANO}{https://github.com/kazizzad/GANO}\\
        \textit{geo-FNO} & \href{https://github.com/neural-operator/Geo-FNO}{https://github.com/neural-operator/Geo-FNO}\\
        \textit{GNO} & \href{https://github.com/neural-operator/graph-pde}{https://github.com/neural-operator/graph-pde}\\
        \textit{MWT-NO} & \href{https://github.com/gaurav71531/mwt-operator}{https://github.com/gaurav71531/mwt-operator}\\
        \textit{PINO} & \href{https://github.com/neural-operator/PINO}{https://github.com/neural-operator/PINO}\\
        \textit{SNO} & \href{https://github.com/vlsf/sno}{https://github.com/vlsf/sno}\\
        \textit{UNO} & \href{https://github.com/ashiq24/UNO}{https://github.com/ashiq24/UNO}\\
        
    \end{tabular}}
    \caption{This table provides links to the source code for various neural operator architectures}
    \label{tab:tab_links}
\end{table}

\begin{table}[h]
\centering
\resizebox{8.5cm}{!}{\begin{tabular}{|l|l|}
\hline
\textbf{Term} & \textbf{Meaning}                         \\ \hline
A             & Banach Function Space                    \\ \hline
a(x)             & Input function                           \\ \hline
$\alpha$ & Hyperparameter for neural network \\ \hline
b             & Bias                                     \\ \hline
$\beta$ & Hyperparameter for neural network \\ \hline
C             & Specific heat capacity                   \\ \hline
$\mathcal{C}$             & Cost Function                  \\ \hline
c             & Speed of Sound                           \\ \hline
D             & Domain                                   \\ \hline
$d_i$, $b_i$             & Finite dimensional vectors                                   \\ \hline
$e_{xy}$ & Edge from node x to node y in a graph neural network \\
\hline
$\mathcal{F}$             & Fourier Transform                        \\ \hline
f(x)             & Function                                 \\ \hline
$\mathcal{F}^{-1} $          & Inverse Fourier Transform                \\ \hline
G             & neural operator                          \\ \hline
$G^+$             & Non-Linear map between Function Spaces                          \\ \hline
$g_i$          & Complex Exponential/Chebyshev polynomial \\ \hline
$h_i$          & Hidden layer Embedding \\ \hline
k             & directional Thermal Conductivity         \\ \hline
K, $\kappa$             & Kernel                                   \\ \hline
L             & Banach Function Space                    \\ \hline
$\mathbf{L}$ & Linear Differential Operator \\ \hline
$\mathcal{L}$             & Loss Function                    \\ \hline
$\lambda$ & Directional Thermal Conductivity \\ \hline
N             & Natural Numbers                    \\ \hline
n & Arbitrary Natural number\\ \hline
P             & Lifting Map                              \\ \hline
p(r,t) & photoacoustic pressure wave at position r, time t \\ \hline
$\mathcal{P}$             & Partial Differential Operator            \\ \hline
$\Phi$ & flux \\ \hline
$\varphi$ & Porosity \\ \hline
$\mathcal{Q}$             & internal heat source                     \\ \hline
Q             & Projecting Map                           \\ \hline
R             & Non-Linear Partial Differential Operator \\ \hline
$\Real$             & Real Numbers \\ \hline
$\rho$           & density                                  \\ \hline
$\varsigma$ & stress tensor \\
\hline
$S_p$          & Saturation of phase p                    \\ \hline
$\sigma$         & Non-Linear Activation                    \\ \hline
$\mathbf{T}$            & temperature                              \\ \hline
t, T             & time                                     \\ \hline

U             & Banach Function Space                    \\ \hline
u(x)             & solution function                        \\ \hline
$\mathcal{U}$ & Velocity \\
\hline
$u_\theta$      & neural network parameterized by $\theta$                           \\ \hline
$v_i$, $v^{(i)}$          & ith output of a neural network                               \\ \hline
$\mathbf{V}$             & neural network Approximation of Solution               \\ \hline
$\nu$             & Viscosity coefficient               \\ \hline
W             & Weight                                   \\ \hline
$\mathcal{W}$ & Weiner Process \\ \hline
w             & Vorticity                                   \\ \hline
X             & mass fraction                            \\ \hline
$\xi$ & noise \\ \hline
$x_i$            & ith input element                        \\ \hline
$\nabla$            & Gradient Operator                      \\ \hline
$\Delta$            & Laplacian Operator                        \\ \hline
\end{tabular}}

\caption{A summary of mathematical notations used in the article}
\label{tab:terms}
\end{table}


%

\end{document}